\PassOptionsToPackage{numbers}{natbib}
\documentclass{jfm}
\usepackage{graphicx}
\usepackage{epstopdf, epsfig}
\usepackage{mdframed}
\usepackage{enumitem}
\usepackage[utf8]{inputenc}
\usepackage[english]{babel}
\usepackage[numbers]{natbib}
\usepackage{hyperref}
\usepackage{{booktabs}}
\usepackage{wrapfig}
\usepackage{float} 
\usepackage{placeins}
\usepackage{graphicx}
\usepackage{caption}
\usepackage[usenames, dvipsnames, svgnames, table]{xcolor}
\usepackage{multirow}
\usepackage[strict]{changepage}

\definecolor{customgreen}{HTML}{009900}
\definecolor{customred}{HTML}{FF0000}
\newcommand{\negative}[1]{\textcolor{customgreen}{#1}}
\newcommand{\positive}[1]{\textcolor{customred}{#1}}

\newmdenv[
  backgroundcolor=gray!10, % adjust the shade of grey as needed
  linewidth=0pt, % no border
  innerleftmargin=10pt,
  innerrightmargin=10pt,
  innertopmargin=10pt,
  innerbottommargin=10pt
]{codeblock}

\title{The Sound of Healthcare: Improving Medical Transcription ASR Accuracy\\ with Large Language Models}

\author{Ayo Adedeji\aff{1},
  Sarita Joshi\aff{1},
  \and Brendan Doohan\aff{1}}

\affiliation{\aff{1}Google Cloud, 1600 Amphitheatre Parkway, Mountain View, CA 94043, USA}

\begin{document}

\maketitle
% \vspace{-0.2em}
% Abstract
\begin{mdframed}[backgroundcolor=gray!20,  linewidth=0pt]
\begin{abstract}
In the rapidly evolving landscape of medical documentation, transcribing clinical dialogues accurately is increasingly paramount. This study explores the potential of Large Language Models (LLMs) to enhance the accuracy of Automatic Speech Recognition (ASR) systems in medical transcription. Utilizing the PriMock57 dataset, which encompasses a diverse range of primary care consultations, we apply advanced LLMs to refine ASR-generated transcripts. Our research is multifaceted, focusing on improvements in general Word Error Rate (WER), Medical Concept WER (MC-WER) for the accurate transcription of essential medical terms, and speaker diarization accuracy. Additionally, we assess the role of LLM post-processing in improving semantic textual similarity, thereby preserving the contextual integrity of clinical dialogues. Through a series of experiments, we compare the efficacy of zero-shot and Chain-of-Thought (CoT) prompting techniques in enhancing diarization and correction accuracy. Our findings demonstrate that LLMs, particularly through CoT prompting, not only improve the diarization accuracy of existing ASR systems but also achieve state-of-the-art performance in this domain. This improvement extends to more accurately capturing medical concepts and enhancing the overall semantic coherence of the transcribed dialogues. These findings illustrate the dual role of LLMs in augmenting ASR outputs and independently excelling in transcription tasks, holding significant promise for transforming medical ASR systems and leading to more accurate and reliable patient records in healthcare settings.
\end{abstract}
\end{mdframed}
\vspace{1em}
% Keywords
\vspace{-3.5em}
% Introduction
\section{Introduction}\label{sec:introduction}
In the nuanced and high-stakes realm of healthcare, the accuracy of medical transcriptions is not just a matter of administrative record but a cornerstone of effective patient care and treatment planning \citep{Zhang2023, Johnson2014}. Automatic Speech Recognition (ASR) systems, now widely used in clinical settings, are expected to reliably and accurately convert spoken interactions into written medical records \citep{Johnson2014, Saxena2017}. However, these systems, despite their advancements, continue to grapple with substantial challenges. They struggle to accurately capture the complexities of medical conversations, which include diverse accents, dialects, and the specialized lexicon of healthcare \citep{Zhou2018, DiChristofano2023}. This difficulty is compounded by the subtlety of speech in clinical settings, where capturing terminologies and nuanced expressions precisely is crucial. 

As such, the errors encountered in medical ASR systems are wide and varied. These include misinterpretations of drug names and dosages, incorrect lab values, left/right anatomical discrepancies, medical inconsistencies, age and gender mismatches, incorrect doctor names, and wrong dates \citep{Hodgson2015}. Further compounding the issue are problems like the introduction of made-up words, along with omissions and duplications \citep{Hodgson2015, McGurk2008}. These inaccuracies in medical transcription can lead to alarming consequences, potentially impacting patient diagnoses and treatment decisions \citep{Adane2019}. Addressing these challenges necessitates innovative approaches that transcend the current capabilities of ASR systems.

In response to these challenges, Large Language Models (LLMs) have emerged as a significant development in the field of artificial intelligence and natural language processing. Developed through extensive training on vast text corpuses, these models have demonstrated a remarkable ability to parse human language and contextualize and interpret it with high accuracy \citep{Naveed2023, Kojima2023}. Recently, some studies have explored the possibility of equipping LLMs with an audio encoder for direct speech recognition, pushing the boundaries of how these models can be applied in the realm of ASR \citep{Fathullah2023, Hono2023, Rubenstein2023}. Conversely, others have experimented with an alternative approach, transferring and incorporating multi-layer representations from LLMs into ASR transducer models to augment their performance \citep{Udagawa2023, Park2023}.

Our study adopts a contrasting approach to the existing advancements in LLM application within ASR. Rather than augmenting LLMs with additional audio processing capabilities, we delve into their inherent text-processing abilities. This approach is grounded in the understanding that text, as a derivative of speech, possesses inherent characteristics which LLMs can effectively utilize without the need for an appended audio encoder. This direction resonates with prior studies that have utilized off-the-shelf or LoRA fine-tuned LLMs to leverage ASR-generated word level hypotheses to improve keyword spotting and intent classification \citep{Dighe2023, Ma2023}. Building on this concept, our approach utilizes LLMs to generate a range of word and sentence level hypotheses intrinsically \citep{Freitag2017}. This method not only facilitates the correction of transcription errors and the resolution of semantically infeasible statements but also accurate speaker diarization by interpreting the nuanced tones within the dialogues.

Our methodology extends beyond word-for-word transcription but aims to capture the essence of medical dialogues with accuracy, effectiveness, and \textit{empathy}. Our research holds particular relevance for the vast quantities of medical transcriptions that exist without accompanying audio, often created using older ASR technologies. In low-income countries, where advanced on-device LLMs may not be practical, our research supports a viable solution: combining affordable audio equipment with LLM post-correction via an API. Our goal is to test the boundaries of ASR and LLM capabilities without extensive modifications, allowing for their application in a wide range of scenarios and tasks. This approach seeks to bridge the technological gap, enhancing the practical utility of these models in diverse healthcare environments.

% Materials
\section{Materials}\label{sec:materials}
% Materials - Dataset
\subsection{Dataset}
Our study utilized the PriMock57 dataset, developed by Babylon Health, comprising 57 mock consultations and encompassing approximately 9 hours of recorded dialogue \citep{Korfiatis2022}. These consultations reflect the diverse range of medical situations commonly encountered in clinical practice, spanning a variety of conditions including cardiovascular issues, dermatological problems, physical injuries, and mental health episodes among others. On average, each consultation has 92 conversation turns and 1,489 spoken words. The dataset features a balanced gender distribution among clinicians and participant actors, with a majority in the age range of 25 to 45 years old. The dataset also showcases a wide range of accents: clinicians mainly speak in British English, while patient roles include Indian, various European, and other accents, approximating the linguistic diversity typical in clinical contexts encountered in the United Kingdom.

To more closely align with real-world clinical interactions, for each consultation, we combined the originally separate doctor and patient audio channels into unified tracks for analysis. This step was important to authentically replicate the dynamics of medical dialogues and to more practically assess the utility of our research in noisy healthcare environments, which are known to adversely affect ASR performance \citep{Quiroz2019, Vogel2009FactorsAT}.

Further enhancing the dataset's utility for our research, the dataset contains ground truth transcripts that were manually curated and diarized word-by-word by experienced human transcribers. This meticulous process provides a reliable standard for evaluating both ASR system performance and LLM enhancements against metrics such as Word Error Rate (WER) and Medical Concept WER (MC-WER).

\begin{figure}
  \centering
  \includegraphics[width=0.85\linewidth, keepaspectratio]{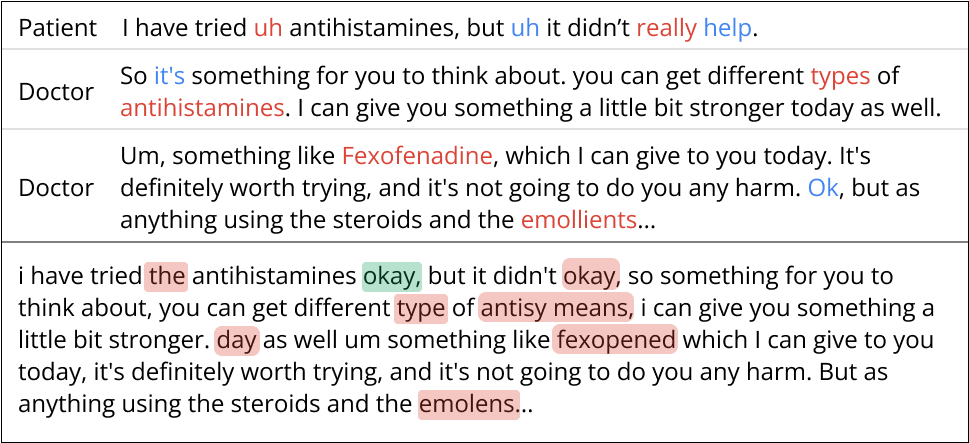}
  \caption{Excerpt from a mock consultation showing the ground truth transcript (top) and the corresponding ASR output (bottom). Errors are annotated: substitutions in red, deletions in blue, and insertions in green.}
  \label{fig:comparing_asr_output_with_ground_truth_transcript}
\end{figure}

% Materials - ASR Systems
\subsection{ASR Systems}
In our analysis, we employed six ASR systems, chosen for their popularity or effectiveness in processing complex, jargon-heavy healthcare language. These systems are:
\vspace{0.1em}
\begin{enumerate}[leftmargin=3em, labelwidth=*, labelsep=.5em, itemindent=0em, align=right, itemsep=0pt, label=\textup{(\roman*)}]
  \item \indent Google Cloud's Medical Conversation model (GCMC): a commercially available model that is specialized for medical conversations \citep{GoogleCloudMedicalConversation}.
  \item Google Cloud's Chirp: a commercially available, universal speech model that is the latest and next generation of Google's speech-to-text models \citep{GoogleCloudChirp}.
  \item OpenAI's Whisper 1: a commercially available, universal speech model \citep{OpenAIWhisper}.
  \item Amazon Transcribe Medical: a commercially available service. We used the conversation model with speciality set to Primary Care \citep{AWSTranscribeMedical}.
  \item Soniox: a commercially available service. We used the en\_v2 model \citep{Soniox}.
  \item Deepgram's Nova 2: a commercially available model. We used the medically tuned version \citep{Deepgram}.
\end{enumerate}
\vspace{1em}

For evaluating the effectiveness of LLM enhancements, we focused on prompting and iterating upon the outputs of GCMC, Chirp, and Whisper 1 in various experiments. Meanwhile, Amazon Transcribe Medical, Soniox, and Deepgram's Nova 2 were leveraged as comparative benchmarks. This dual approach allowed us to assess both the baseline performance of leading ASR systems and the incremental improvements brought about by LLM post-processing.
% Methods - Large Language Models
\subsection{Large Language Models}
In our analysis, we integrated and tested several Large Language Models (LLMs), selected either for their wide adoption or proven efficacy in natural language understanding and generation, making them ideal candidates for our study's focus:
\vspace{0.5em}
\begin{enumerate}[leftmargin=3em, labelwidth=*, labelsep=.5em, itemindent=0em, align=right, itemsep=0pt, label=\textup{(\roman*)}]
  \item Google Cloud’s Gemini Pro: a general purpose model efficiently optimized for tasks like brainstorming, summarizing content, and writing \citep{GoogleCloudGemini}.
  \item Google Cloud’s Gemini Ultra: a general purpose model that is Google’s largest and most capable model for highly complex tasks \citep{GoogleCloudGemini}.
  \item Google Cloud’s Text Bison 32k: a general purpose model that is the most advanced of Google’s prominent Bison line \citep{GoogleCloudVertexAI}.
  \item Anthropic’s Claude V2: a general purpose model that is optimized for dialogue, content creation, complex reasoning, creativity, and coding \citep{AnthropicClaude2}.
  \item OpenAI’s GPT-4: a general purpose model that is optimized for complex instructions in natural language and can solve difficult problems with accuracy \citep{OpenAIGPT4}.
  \item Meta’s LLaMA 2: a general purpose, open-source model pretrained on publicly available online data sources that is optimized for dialogue use cases. We used the Chat 70B model \citep{MetaLLama2}.
\end{enumerate}
\vspace{1em}

In our experiments, all LLMs were configured to operate at their maximum output token limit, ensuring the widest possible range of language generation. Additionally, we set the models to a temperature setting between 0.1 and 0.2.  This calibration was designed to reduce variability in responses, thereby aligning the models' outputs more closely with a greedy decoding strategy \citep{Freitag2017, Chen2018}. This choice was particularly advantageous given the typical error characteristics of medical ASR systems, which are generally accurate but prone to specific types of errors such as word substitutions and omissions \citep{Hodgson2015, McGurk2008}. A smaller beam width mitigates the risk of introducing unnecessary changes and over-correcting \citep{Jinnai2023}. Such a focused approach becomes even more relevant when the model’s training data might not fully align with the specific language patterns of the transcripts being corrected. In prioritizing fewer, more targeted modifications, our approach was designed to preserve the integrity of the original ASR output, focusing on correcting the most apparent errors.
% Materials - Embedding Models
\subsection{Embedding Models}
In the evaluation of semantic similarity between ground truth transcripts, ASR-generated transcripts, and LLM enhanced outputs, we employed a range of widely adopted and advanced embedding models:
\vspace{0.5em}
\begin{enumerate}[leftmargin=3em, labelwidth=*, labelsep=.5em, itemindent=0em, align=right, itemsep=0pt, label=\textup{(\roman*)}]
  \item Google Cloud’s PaLM Gecko: a smaller, derivative version of the PaLM 2 model. It accepts a maximum of 3,072 input tokens and outputs 768-dimensional vector embeddings \citep{GoogleCloudTextEmbeddings}.
  \item BERT: a bidirectional transformer pretrained. It accepts a maximum of 512 input tokens and outputs 768-dimensional vector embeddings \citep{Devlin2019}.
  \item RoBERTa: a derivative of BERT, RoBERTa modifies the BERT pretraining procedure by training on a larger dataset and with larger mini-batches and learning rates. It also does not have BERT’s next sentence prediction objective. It accepts a maximum of 512 input tokens and outputs 768-dimensional vector embeddings \citep{Liu2019}.
  \item OpenAI’s Ada: a commercial, general embedding model. It accepts a maximum of 8,192 input tokens and outputs 1,536-dimensional vector embeddings \citep{OpenAIEmbeddings}.
\end{enumerate}
\vspace{1em}

\begin{figure}
  \centering
  \includegraphics[height=0.32\textheight,keepaspectratio]{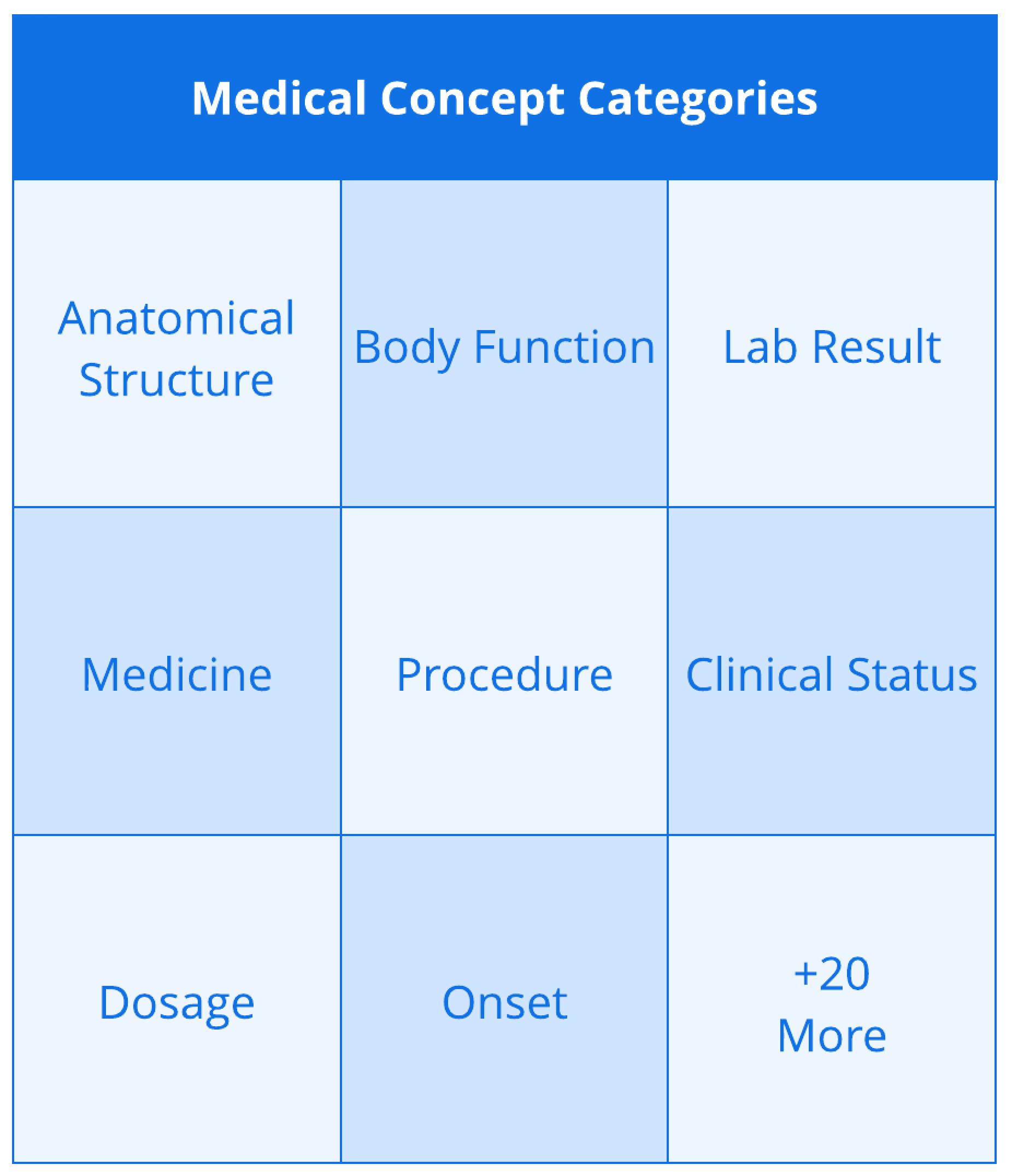}
  \caption{Visual of medical concepts recognized by the Healthcare Natural Language API}
  \label{fig:medical_concept_categories}
\end{figure}

A common challenge in applying these models lies in their input token constraints. For instance, BERT and RoBERTa, without fine tuning, are limited to texts no longer than 512 tokens \citep{Jaiswal2023}. To address this, we segmented the ASR and LLM outputs line by line. We then applied cosine similarity matching on these segmented lines. The similarity scores of all lines were subsequently averaged to provide an aggregate measure of semantic similarity. This approach allowed us to circumvent the token limitation and conduct an accurate semantic analysis across the entirety of the transcribed outputs.
% Materials - Cloud Healthcare Natural Language
\subsection{Cloud Healthcare Natural Language}
Lastly, our study employed Google Cloud's Healthcare Natural Language API to contextually and semantically annotate medical terms within the transcriptions \citep{GoogleCloudHealthcareNLP}. This API is specifically engineered to parse and structure unstructured medical text, such as medical records or insurance claims. Key functionalities of the Healthcare Natural Language API include:
\vspace{0.5em}
\begin{enumerate}[leftmargin=3em, labelwidth=*, labelsep=.5em, itemindent=0em, align=right, itemsep=0pt, label=\textup{(\roman*)}]
  \item Medical Concept Extraction: It identifies and extracts medical concepts such as diseases, medications, and procedures, mapping these concepts to standard medical vocabularies like RxNorm, ICD-10, and SNOMED CT.
  \item Medical Concept Categorization: The API categorizes medical concepts into over thirty various entities like anatomical structure, medicine, procedure, and laboratory data, facilitating a detailed understanding of medical terminologies.
  \item Functional Features Analysis: It assesses functional features such as temporal relationships, subjects, and certainty assessments of medical conditions and treatments, enhancing the depth of medical data interpretation.
\end{enumerate}
\vspace{1em}

The selection of this API was underpinned by its well-tested, leading recall and accuracy in extracting and annotating complex healthcare information \citep{McKnightDolezal2022}. Its ability to map medical concepts to established medical vocabularies and to provide a structured representation of the medical entities made it a key tool in the computation of MC-WER. The API's support for a diverse range of medical vocabularies and knowledge categories is particularly well-suited to the PriMock57 dataset's varied medical dialogues and consultation types. This compatibility enabled our analysis to comprehensively encompass a broad spectrum of medical terminology, effectively addressing both commonplace and rare medical conditions. Such versatility is essential for accurately reflecting the nuanced and diverse nature of real-world clinical conversations.
% Methods
\section{Methods}\label{sec:methods}
% Methods - Zero-shot Prompting
\subsection{Zero-shot Prompting}
Zero-shot prompting was utilized to determine the LLMs' proficiency in diarizing and correcting ASR outputs in a single inference step. Zero-shot prompting is a technique where an LLM is presented with a task and an instructional prompt without prior task-specific examples \citep{Kojima2023}. It leverages the model's pretrained knowledge and its ability to infer context from the prompt alone to generate an accurate response.
\begin{wrapfigure}{r}{0.5\textwidth} % "r" for right and "0.5\textwidth" for the width of the figure
  \centering
  \includegraphics[height=0.30\textheight,keepaspectratio]{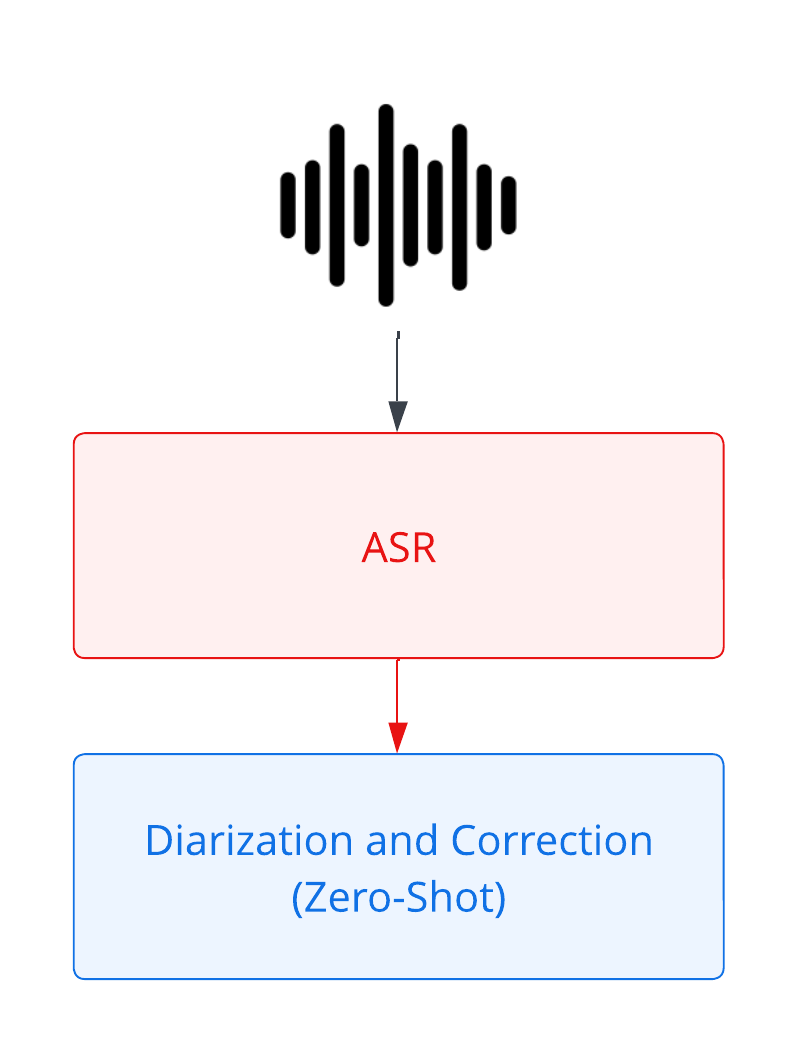}
  \caption{Workflow of Zero-Shot Prompting for ASR Diarization and Correction.}
  \label{fig:zero_shot_prompting}
\end{wrapfigure}

We implemented zero-shot prompting in a structured manner, engaging with transcript segments that varied in length. Our systematic approach included concise snippets of dialogue, in both 5 and 10-line chunks, as well as processing complete transcripts in a single iteration. This variability allowed us to assess the LLMs' zero-shot performance across different context window sizes, which has been noted to negatively influence task performance with larger windows \citep{Liu2023lost, Chen2023longlora}. We additionally cast the LLM in the role of a speech-to-text transcription assistant. This role-playing approach has been noted to impact task performance positively \citep{Kong2023better}. Our methodology can be formalized as follows:

\begin{equation}
y = LLM(I, x)
\end{equation}\\
Here, x represents the ASR-generated transcript segments provided to the LLM, and y is the LLM's output, which should be the corrected and diarized transcript. The variable I is the instructional prompt that directs the LLM to contextualize and process the ASR-generated transcript segments.\\

For this task, the following prompt template was utilized:\\

\begin{codeblock}
You are a helpful speech-to-text transcription assistant. Your task is to review and correct transcription errors, focusing on accuracy and context. Consider diverse speaker accents. Identify the role of each speaker, either a doctor or patient, based on tone, sentiment, and diction. Label them accordingly in the transcript. Ensure the enhanced text mirrors the original spoken content without adding new material. Your goal is to create a transcript that is accurate and contextually consistent, which will improve semantic clarity and reduce word error rates.\\

\noindent Transcript:\\
\{transcript\_segments\}\\

\noindent Enumerated and enhanced transcript:\\
\#. Speaker \(Patient/Doctor\): \string[enhanced sentence\string]
\end{codeblock}
\vspace{0.1em}
% Methods - Chain-of-Thought Prompting
\subsection{Chain-of-Thought Prompting}
We also explored a Chain-of-Thought (CoT) prompting strategy designed to enhance the LLMs' performance. CoT prompting has been shown to significantly improve performance on complex reasoning tasks \citep{Wei2023chainofthought, Wu2023analyzing}. CoT prompting enriches the LLM's problem-solving process by providing a series of intermediate reasoning steps accompanied by few-shot examples \citep{Madaan2022text, Wang2023understanding}. This approach guides the model's reasoning and shows how to apply context and intermediate rationale to arrive at an accurate answer.
\begin{wrapfigure}{r}{0.5\textwidth} % "r" for right and "0.5\textwidth" for the width of the figure
  \centering
  \includegraphics[height=0.35\textheight,keepaspectratio]{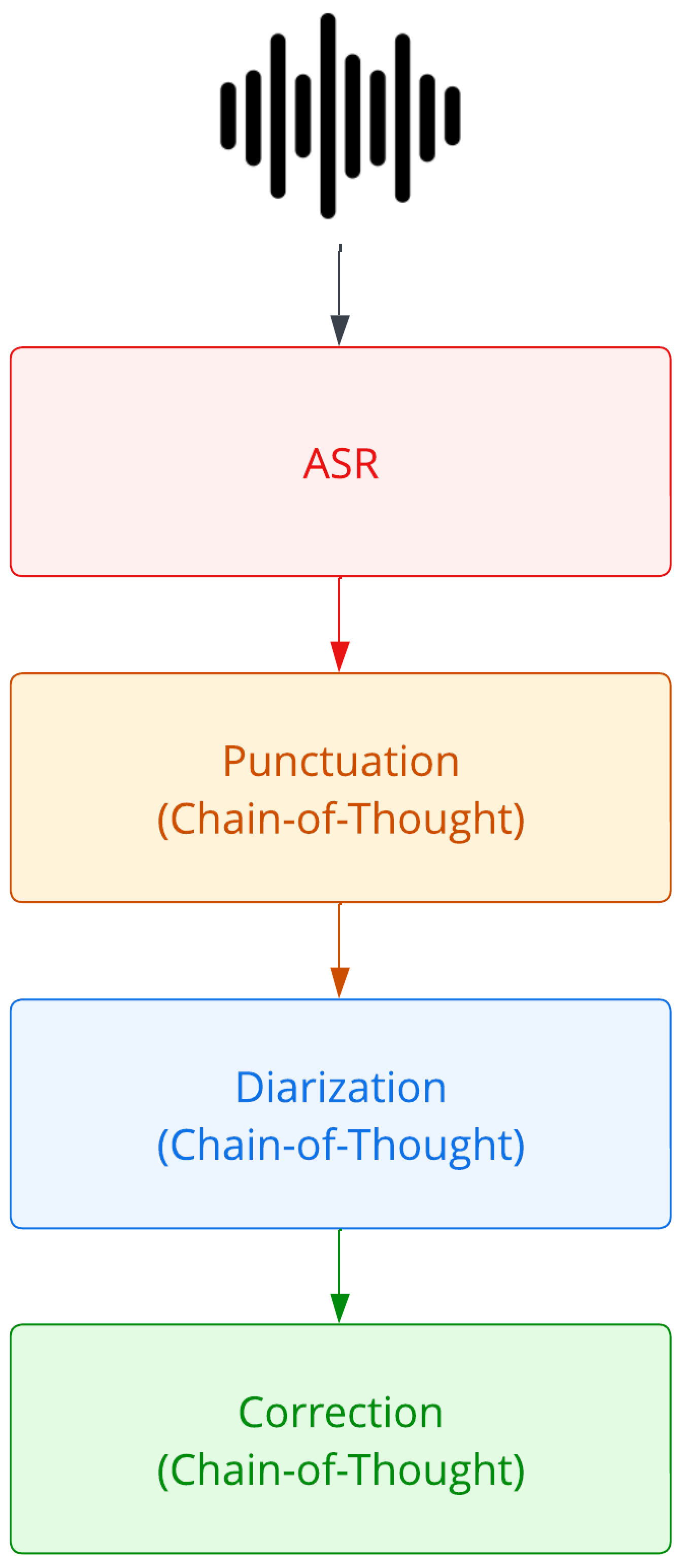}
  \caption{Workflow of Chain-of-Thought Prompting for ASR Punctuation, Diarization and Correction.}
  \label{fig:chain_of_thought_prompting}
\end{wrapfigure}

In our methodology, diarization and correction were separated into distinct inference steps. This separation aligns with evidence suggesting that LLMs achieve better performance when addressing individual components of a multifaceted task sequentially, rather than concurrently \citep{Shi2023large, Chen2023program, Pandia2021sorting}. Additionally, to counteract the adversarial impact of insufficient or incorrect punctuation on diarization accuracy—a subject we explore in depth in the Results section—an initial step focusing on punctuation enhancement was incorporated into our CoT prompting strategy. This initial action serves to refine the input text, providing the LLM with a more coherent and structured foundation for the ensuing diarization and correction tasks. Our CoT prompting strategy at each punctuation, diarization, and correction step can be formalized as follows:
\vspace{0.2em}
\begin{equation}
y = LLM(I, (x_1, r_1, y_1), (x_2, r_2, y_2), ..., (x_k, r_k, y_k), x)
\end{equation}
\vspace{0.1em}

Here, x denotes the input transcript at each stage. Initially, x is the raw ASR-generated transcript for the punctuation enhancement stage. Subsequently, x becomes the punctuated output for the diarization stage and the diarized output for the final correction stage. The pairs ($x_1$, $r_1$, $y_1$),  ($x_2$, $r_2$, $y_2$), . . .,($x_k$, $r_k$, $y_k$) represent the few-shot examples with their corresponding rationales and expected outputs, which are provided to the LLM to facilitate the CoT process. The variable y is the LLM’s output, which is contextually defined based on the stage: it is an intermediate output post-punctuation and post-diarization, and it becomes the final corrected transcript after the correction stage.

In a parallel experiment, we omitted the punctuation and diarization steps and directly passed ASR-generated transcript segments to the CoT correction prompt. This allowed us to compare and contrast the incremental advantages provided by an iterative CoT LLM post-processing approach over that of a single-step correction.\\

For this task, the following is an exemplar prompt template for diarization:\\
% Diarization Prompt Template
\begin{codeblock}
You are a helpful speech-to-text transcription assistant. Your current task is to diarize a conversation with no speaker labels. You will use your advanced understanding of medical terminology, dialogue structure, and conversational context to diarize the text accurately. Here's how to approach the task step by step:\\
1. Contextual Reading: Read each sentence thoroughly, absorbing its content, tone, sentiment, and vocabulary.\\
2. Sentence Splitting: Actively split sentences into separate statements when there's a change in speaker. Look for cues like pauses, speech direction changes, thought conclusions, questions, and answers.\\
3. Reasoning: Consider whether the language is technical (suggesting a medical professional) or expresses personal experiences/emotions (suggesting the patient).\\
4. Look-Around Strategy: Analyze the five sentences before and after the current one to understand the conversation flow. Questions may be followed by answers, and concerns by reassurance.\\
5. Label with Justification: Assign a label 'Doctor' or 'Patient' to each sentence, providing a brief justification based on your analysis. Ensure each justification pertains to only one person.\\
6. Consistent Attribution: Maintain a thorough approach throughout the transcript, treating each sentence with equal attention to detail.\\
7. Extremely Granular Attribution: Break down the conversation into the smallest parts (question, answer, utterance) for clarity. Each clause should be precisely attributed to either the doctor or the patient, with no overlap in speaker identity.\\
8. Examples \string#1-5:\\
\{examples\}\\

\noindent Transcript:\\
\{transcript\_segments\}\\

\noindent Expected enumerated output structure:\\

\noindent Sentence \string#: \string[reference sentence\string]\\
Justification: \string[justification\string]\\
Label: Speaker (Whoever is most likely between "Doctor" and "Patient")
\end{codeblock}
% Methods - Regex Parsing
\subsection{Regex Parsing}
In our study, we incorporated an extensive regex-based post-processing strategy to extract and optimize the integrity of outputs from the LLMs. Given the unique output tendencies of each LLM, deviations from our prompted expected output formats were not uncommon. These deviations typically manifested as missing components like rationales, unintended prose at the beginning or end of outputs, and variations in the use of brackets or parentheses.\\

The following are simplified examples of the regex expressions used in our analysis:\\

\indent For extracting punctuated sentences:\\
\begin{codeblock}
re.compile(r"Punctuated Sentence\string\s*:.*?")
\end{codeblock}
\vspace{4mm}

For extracting diarized sentences:\\
\begin{codeblock}
re.compile(r"Label:\string\s*Speaker\string\s*\((.*?)\)")
\end{codeblock}
\vspace{4mm}

For extracting corrected sentences:\\
\begin{codeblock}
re.compile(r"Corrected Sentence:\string\s*(.*?)\string\s*")
\end{codeblock}
\vspace{4mm}

We additionally employed the Smith-Waterman alignment method in our post-processing strategy to address the challenge of text degeneration, a common issue in outputs generated through greedy decoding and low beam widths \citep{Holtzman2020curious, Chiang2021relating}. This alignment process compared the regex-parsed LLM outputs with their corresponding inputs, which were either the original segments of ASR-generated transcripts or intermediate outputs previously generated by the LLMs. We focused primarily on identifying unaligned words at the end of LLM outputs. We established a threshold of 20 unaligned words: any LLM output with more than 20 unaligned words at its end compared to its input was indicative of text degeneration. To optimize the integrity and relevance of the LLM output for the computation of WER and MC-WER, we truncated these excessive unaligned portions.
% Methods - Word Error Rate
\subsection{Word Error Rate}
Word Error Rate (WER) is a standard measure used to assess the accuracy of ASR systems by comparing the transcribed, hypothesis text against a reference transcript.\\

The WER is calculated using the formula:\\

\begin{equation}
WER = \frac{S + D + I}{N}
\end{equation}\\
\noindent , where S is the number of substitutions, D is the number of deletions, I is the number of insertions, and N is the number of words in the reference transcript.\\

To ensure a fair and consistent evaluation across all ASR outputs, we standardized the transcripts through the following post-processing steps:

\vspace{0.5em}
\begin{enumerate}[leftmargin=3em, labelwidth=*, labelsep=.5em, itemindent=0em, align=right, itemsep=0pt, label=\textup{(\roman*)}]
  \item Removal of Disfluencies: Disfluencies such as "umm", "uhh", and "ahh" were removed from both hypothesis and reference transcripts.
  \item Numeral Normalization: Numerals (e.g., "6", "89") were converted to their written forms ("six", "eighty-nine") for consistency.
  \item Punctuation and Case Normalization: All punctuation was removed, spaces were standardized, and all text was converted to lowercase.
  \item Spelling Normalization: Variations in British and American English spelling were normalized.
  \item Hyphenation Normalization: Hyphenated words were split into two separate words to accommodate the varied handling of hyphenation by different ASR systems.
\end{enumerate}

\vspace{1em}
For the computation of WER, we utilized the Jiwer Python library \citep{Jiwer}.
\vspace{1em}

And relatedly, to evaluate diarization accuracy, we compared dialogue attributed to a specific speaker (either by ASR or LLM) with corresponding labeled dialogue in the reference transcript. This approach highlighted errors such as deletions or insertions from misattributed words while simultaneously providing insight into speaker-specific transcription accuracy, which encompasses aspects like substitutions. To sharpen this assessment, salutations like 'hello', 'good afternoon', and 'goodbye'—typically situated at the beginning and end of dialogues—were excluded specifically from the diarization analysis since the relative order of who said “hello” or “goodbye” first and their attribution was not considered significant for our analysis. This exclusion allowed us to focus and measure accuracy on substantive, medically relevant dialogue not obscured by routine filler statements.

% Methods - Medical Concept WER
\subsection{Medical Concept WER}

\begin{figure}
  \centering
  \includegraphics[width=0.88\linewidth]{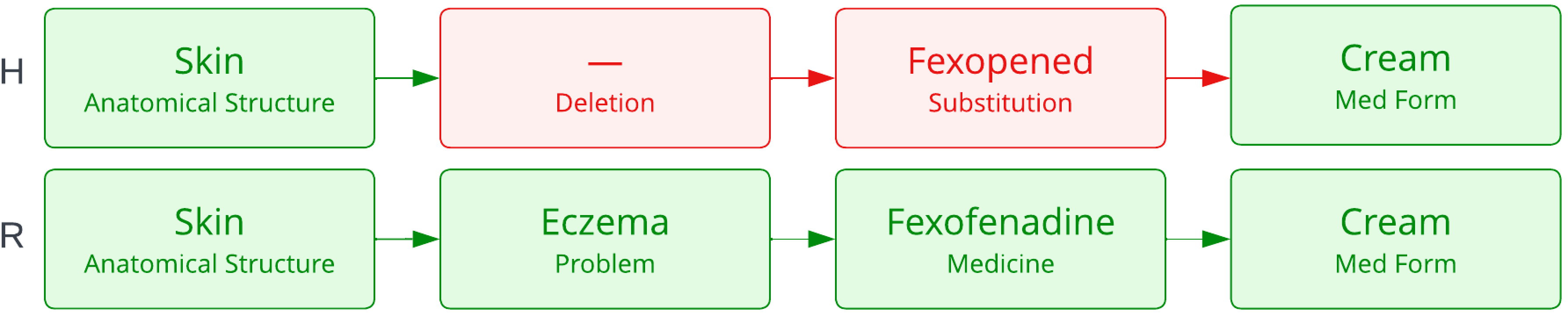}
  \caption{Illustration of MC-WER, comparing hypothesis (H) and reference (R) transcripts. Discrepancies are highlighted in red, demonstrating a deletion and a substitution error.}
  \label{fig:medical_concept_wer}
\end{figure}

In our study, we also evaluated the Medical Concept Word Error Rate (MC-WER), a metric distinct from the general Word Error Rate (WER). While WER treats all words equally, regardless of their significance, MC-WER specifically zeroes in on the accuracy of medical terms in transcription. This distinction is critical in healthcare settings, where precise transcription of terms like 'fexofenadine' as opposed to common words like 'brunch' can have significant implications.\\

To compute MC-WER, we performed the following steps:
\vspace{0.5em}
\begin{enumerate}[leftmargin=3em, labelwidth=*, labelsep=.5em, itemindent=0em, align=right, itemsep=0pt, label=\textup{(\roman*)}]
    \item Normalization: We applied standard normalization procedures to both hypothesis (ASR and LLM outputs) and reference transcripts, including disfluency removal, number conversion, and standardization of punctuation, case, and spelling.
    \item Healthcare NLP Entity Annotation: Using Google Cloud's Healthcare Natural Language API, we annotated medical terms in both the hypothesis and reference transcripts. This extensive annotation process enabled accurate identification and categorization of medical concepts.
    \item Alignment and Error Identification: The Smith-Waterman alignment method was employed to align medical terms between hypothesis and reference transcripts. This process brought to light various error types, such as medical term substitutions ('Dioralyte' incorrectly transcribed as 'diuretics') and instances where general or nonsensical ASR outputs corresponded to medical terms ('teaching' with 'itching', 'airbows' with 'elbows'). These examples from pre-corrected ASR outputs highlight common error patterns and the complexities inherent in ASR systems in medical contexts.
\end{enumerate}
\vspace{4mm}

Our approach in calculating MC-WER, through NLP entity annotation and detailed error analysis, ensured an in-depth evaluation of LLMs’ performance in medical contexts. A more comprehensive analysis of these medical concept errors is presented in the Results
section.

% \vspace{-14em}
\section{Results}\label{sec:results}
\subsection{Diarization}
In evaluating diarization accuracy, we compared the performance of LLMs with that of ASR systems—Soniox, Deepgram Nova 2, and GCMC—selected for their native diarization features. CoT iteration was performed against Chirp's, Whisper's, and GCMC's outputs both in discrete and  aggregated 10-line chunks and in their entirety all at once to assess the effectiveness of LLMs across varying input context window sizes.

Given Chirp and Whisper do not natively possess diarization features, this provided a blank slate for LLMs to demonstrate their diarization capabilities without preexisting speaker labels. Conversely, for GCMC's output, which includes native diarization, we intentionally removed ASR-generated speaker labels before CoT iteration. This approach enabled a fair and balanced comparison, positioning LLM-derived diarization directly alongside ASR-derived benchmarks.

Delving into Doctor-Specific diarization within the scope of the 10-line chunk experiments, the best-performing LLM and ASR system combinations, particularly GPT-4, Gemini Pro, and Gemini Ultra paired with Whisper 1, demonstrated superior diarization accuracy compared to each and every ASR benchmark. These pairings yielded a lower Doctor-Specific Word Error Rate (D-WER) and exhibited reduced variability in results, indicating more reliable and consistent diarization performance (as detailed in Table 1 and illustrated in Figure 6).

In contrast, our examination of Patient-Specific diarization within the same experimental conditions painted a slightly different picture (as detailed in Table 2 and illustrated in Figure 7). Although the LLM-ASR combinations did not outperform the leading ASR systems, Soniox and Deepgram Nova 2, in Patient-Specific Word Error Rate (P-WER), they demonstrated competitive levels of accuracy, lagging by a slim margin of less than \textit{one} percentage point. This marginal discrepancy not only underscored the overall competitiveness of LLM-derived diarization but also aligned with its observed potential to exceed the performance benchmarks set by the ASR systems in the diarization task.

\FloatBarrier
\begin{minipage}{\dimexpr\textwidth-0.5cm} % Use the full width of the page
\centering
\fontsize{8pt}{11pt}\selectfont % Set the font size to 9pt and the line spacing to 11pt
\setlength{\tabcolsep}{8pt} % Adjust the value to your preference
\begin{tabular}{| l r r | r |}
\hline
           LLM &                       STT &                   Method &          D-WER \\
\hline
         GPT-4 &                 Whisper 1 &              Diarization &  10.77\% ± 3.32 \\
    Gemini Pro &                 Whisper 1 &              Diarization &  11.27\% ± 4.15 \\
  Gemini Ultra &                 Whisper 1 &              Diarization &  11.39\% ± 3.92 \\
            -- &                    Soniox &                      ASR & 12.15\% ± 11.01 \\
            -- &           Deepgram Nova 2 &                      ASR &  12.31\% ± 8.82 \\
  Gemini Ultra &                     Chirp & Diarization + Correction &  13.97\% ± 3.80 \\
         GPT-4 &                     Chirp &              Diarization &  14.01\% ± 3.90 \\
    Gemini Pro &                     Chirp & Diarization + Correction &  15.40\% ± 4.43 \\
         GPT-4 &                      GCMC &              Diarization &  23.67\% ± 5.75 \\
  Gemini Ultra &                      GCMC & Diarization + Correction &  24.34\% ± 5.73 \\
    Gemini Pro &                      GCMC & Diarization + Correction &  24.64\% ± 6.13 \\
text-bison-32k &                      GCMC &              Diarization &  24.67\% ± 6.12 \\
     Claude V2 &                      GCMC & Diarization + Correction &  24.71\% ± 5.95 \\
            -- &                      GCMC &                      ASR & 39.37\% ± 14.71 \\
\hline
\end{tabular}
\captionof{table}{Optimal Pairings of LLMs and ASR Systems for Doctor-Specific Diarization in experiments processing entire transcripts in 10-line chunks.}
\label{tab:diarization_accuracy_doctor_table-10_chunks}
\vspace{2em}

\includegraphics[width=1\linewidth]{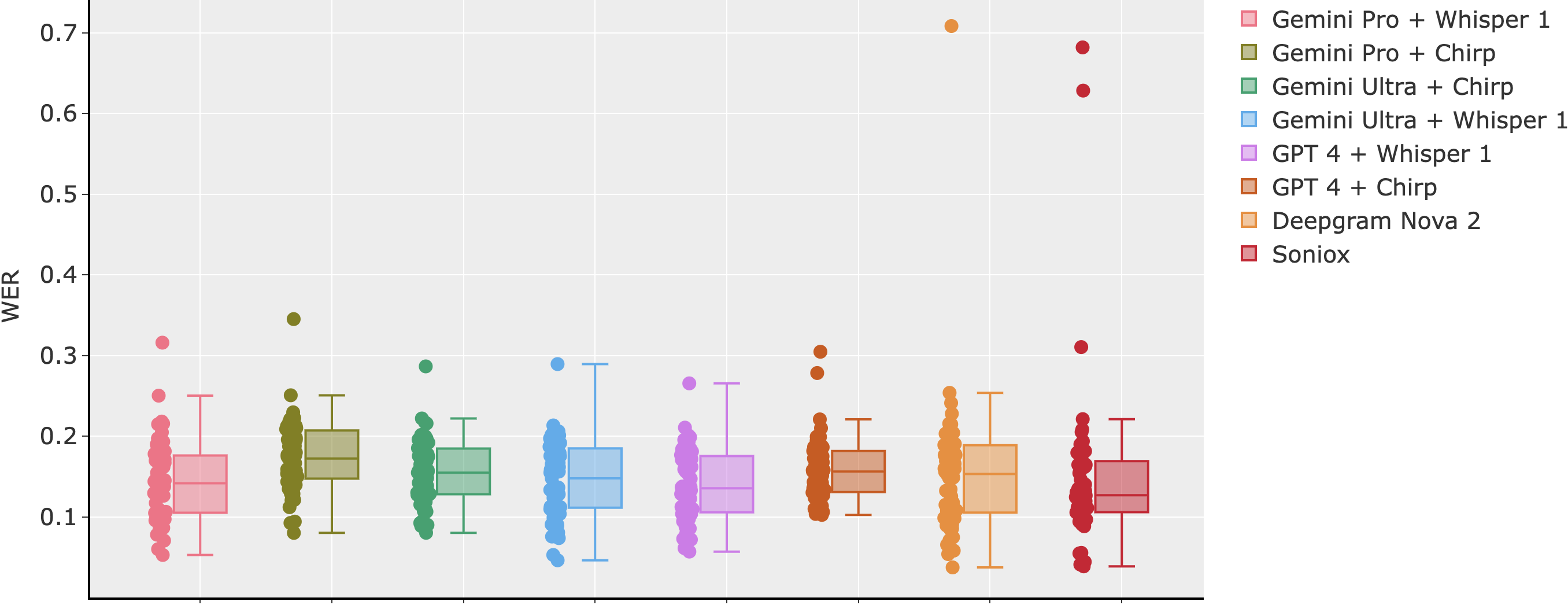} % Adjust the width as needed
\captionof{figure}{Box plot distribution of Doctor-Specific Diarization for top-performing LLM and STT pairings when processing entire transcripts in 10-line chunks.}
\label{fig:diarization_accuracy_doctor_box_plots-10_chunks}
\vspace{2em}
\fontsize{8pt}{11pt}\selectfont % Set the font size to 9pt and the line spacing to 11pt
\setlength{\tabcolsep}{8pt} % Adjust the value to your preference
\begin{tabular}{| l r r | r |}
\hline
           LLM &             STT &                   Method &          P-WER \\
\hline
            -- & Deepgram Nova 2 &                      ASR &  18.02\% ± 6.55 \\
            -- &          Soniox &                      ASR & 18.08\% ± 13.77 \\
         GPT-4 &       Whisper 1 &              Diarization &  18.32\% ± 6.34 \\
    Gemini Pro &       Whisper 1 &              Diarization &  18.92\% ± 6.83 \\
  Gemini Ultra &       Whisper 1 &              Diarization &  19.21\% ± 7.10 \\
  Gemini Ultra &           Chirp &              Diarization &  20.63\% ± 7.15 \\
         GPT-4 &           Chirp &              Diarization &  21.84\% ± 8.38 \\
    Gemini Pro &           Chirp & Diarization + Correction &  22.94\% ± 8.85 \\
         GPT-4 &            GCMC &              Diarization & 32.19\% ± 11.22 \\
text-bison-32k &            GCMC &              Diarization & 33.93\% ± 11.20 \\
    Gemini Pro &            GCMC & Diarization + Correction & 34.06\% ± 12.14 \\
     Claude V2 &            GCMC &              Diarization & 34.27\% ± 12.05 \\
  Gemini Ultra &            GCMC & Diarization + Correction & 34.87\% ± 12.22 \\
            -- &            GCMC &                      ASR & 47.01\% ± 14.95 \\
\hline
\end{tabular}
\captionof{table}{Optimal Pairings of LLMs and ASR Systems for Patient-Specific Diarization in experiments processing entire transcripts in 10-line chunks.}
\label{tab:diarization_accuracy_patient_table-10_chunks}
\end{minipage}

\begin{minipage}{\dimexpr\textwidth}
  \begin{minipage}{\dimexpr\textwidth-0.5cm}
      \centering % Center the figure within the minipage
      \includegraphics[width=1\linewidth]{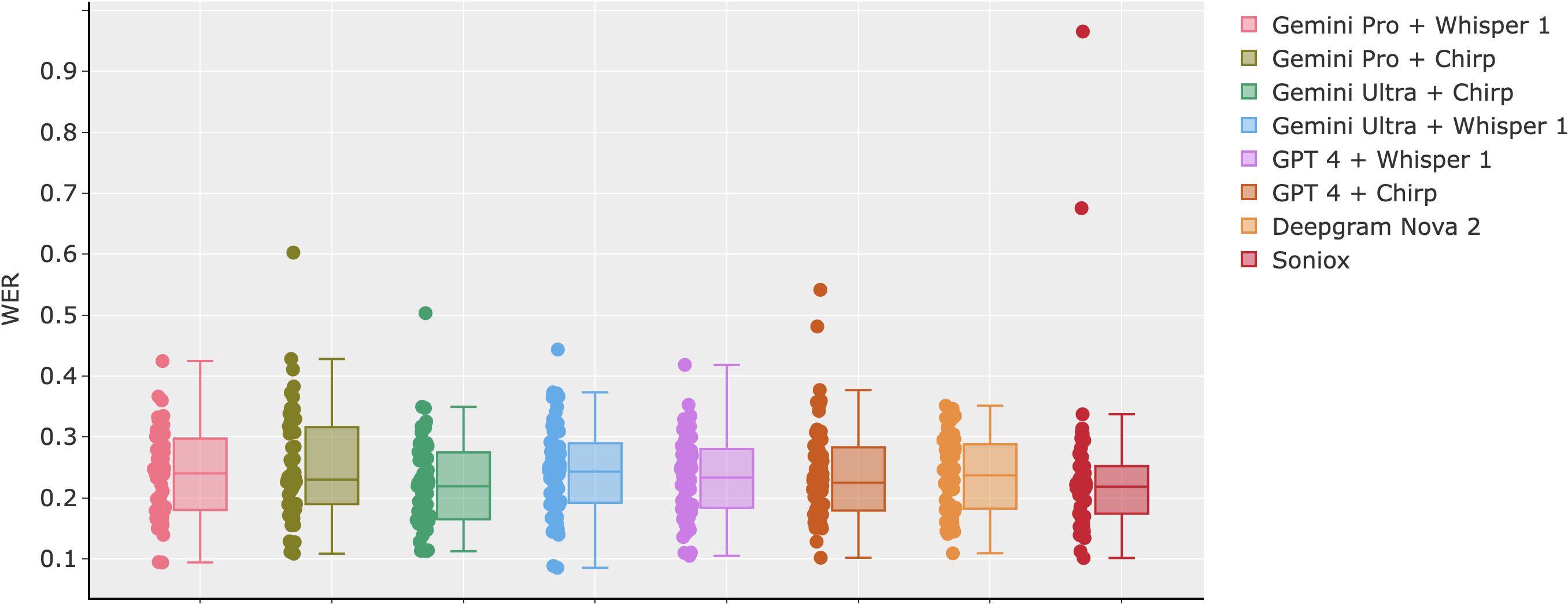} % Adjust the width as needed
      \captionof{figure}{Box plot distribution of Patient-Specific Diarization for top-performing LLM and STT pairings when processing entire transcripts in 10-line chunks.}
      \label{fig:diarization_accuracy_patient_box_plots-10_chunks}
  \end{minipage}
\vspace{2.5em}
\setlength{\parindent}{1em}

Relatedly, in experiments that processed transcripts in their entirety all at once, an LLM-ASR combination did outperform each and every ASR benchmark in Patient-Specific diarization (as detailed in Table 3 and illustrated in Figure 8). The best results for Patient-Specific diarization in these all-at-once experiments were observed with the pairing of Gemini
Pro and Whisper 1, which achieved a cross-experiment P-WER best of 17.90\%. The pairings of Claude V2 and text-bison-32k with Whisper 1 were competitive and lagged the top ASR benchmarks by a slim margin of less than \textit{half} of a percentage point and less than \textit{one} percentage point, respectively. Note, in the all-at-once experiments, Gemini Ultra and GPT-4 were not included due to limitations surrounding their max output token limit and observed inability to output entire transcripts in a single inference step, respectively.\\
\indent Continuing with the analysis of experiments processing transcripts in their entirety all at once, the LLM-ASR combinations, particularly Gemini Pro, text-bison-32k, and Claude V2 with Whisper 1,
replicated their success from the 10-line chunk experiments in Doctor-Specific Diarization, surpassing ASR benchmarks. Notably, the pairing of Gemini Pro with Whisper 1 achieved a cross-experiment D-WER best of 10.04\%, setting a new standard in this category.
\vspace{2em}

\fontsize{8pt}{11pt}\selectfont % Set the font size to 9pt and the line spacing to 11pt
\setlength{\tabcolsep}{8pt} % Adjust the value to your preference
% \vspace{-0.9em}
\begin{tabular}{| l r r | r |}
\hline
           LLM &             STT &                   Method &          P-WER \\
\hline
    Gemini Pro &       Whisper 1 &              Diarization &  17.90\% ± 6.83 \\
            -- & Deepgram Nova 2 &                      ASR &  18.02\% ± 6.55 \\
            -- &          Soniox &                      ASR & 18.08\% ± 13.77 \\
     Claude V2 &       Whisper 1 &              Diarization &  18.43\% ± 7.11 \\
text-bison-32k &       Whisper 1 &              Diarization &  18.68\% ± 7.31 \\
    Gemini Pro &           Chirp & Diarization + Correction &  22.81\% ± 8.31 \\
     Claude V2 &           Chirp &              Diarization & 26.12\% ± 12.45 \\
text-bison-32k &           Chirp & Diarization + Correction & 31.23\% ± 13.96 \\
    Gemini Pro &            GCMC &              Diarization & 32.17\% ± 11.09 \\
text-bison-32k &            GCMC &              Diarization & 36.99\% ± 11.35 \\
            -- &            GCMC &                      ASR & 47.01\% ± 14.95 \\
     Claude V2 &            GCMC & Diarization + Correction & 50.77\% ± 44.42 \\
\hline
\end{tabular}
\captionof{table}{Optimal Pairings of LLMs and ASR Systems for Patient-Specific Diarization in experiments processing entire transcripts all at once.}
\label{tab:diarization_accuracy_patient_table-all_at_once}
\end{minipage}
\FloatBarrier

\begin{minipage}{\dimexpr\textwidth-0.5cm}
  \centering % Center the figure within the minipage
  \includegraphics[width=1\linewidth]{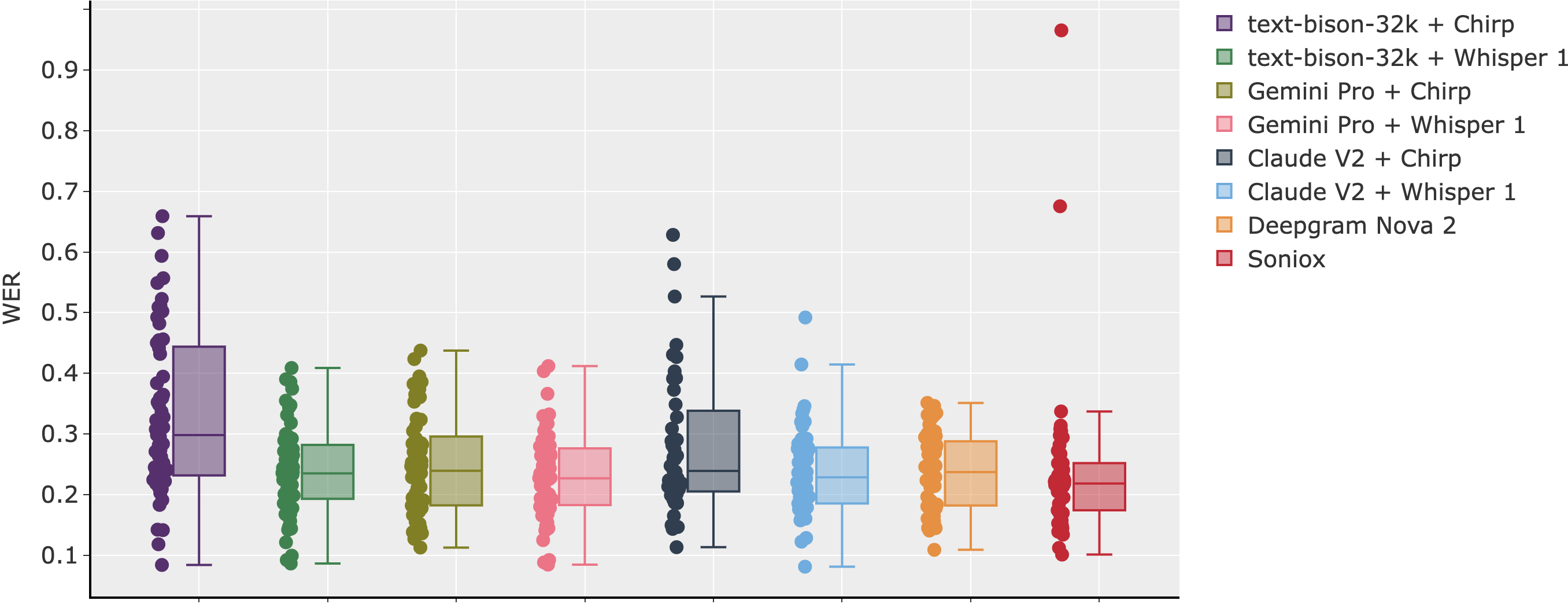} % Adjust the width as needed
  \captionof{figure}{Box plot distribution of Patient-Specific Diarization for top-performing LLM and STT pairings when processing entire transcripts all at once.}
  \label{fig:diarization_accuracy_patient_box_plots-all_at_once}
\end{minipage}
\vspace{1em}

The overall trend of better performance in Doctor-Specific diarization compared to Patient-Specific diarization by an average of 8 basis points, observed in both LLM-ASR combinations and ASR benchmarks, mirrors the linguistic dynamics of our dataset, as previously detailed in the Materials section. Given that clinicians primarily speak in British English, contrasted with the more varied accents of patient roles, this corroborates the challenges we outlined in our introduction, particularly the struggles ASR systems face in transcribing a diverse array of accents and dialects effectively. The improvement conferred by LLM-ASR combinations suggest an enhanced capability in managing linguistic diversity, albeit constrained due to inherent limitations in current ASR systems' adaptability to a wide range of accents and dialects.
\subsection{Adversarial Punctuation}
\begin{minipage}{\dimexpr\textwidth-0.5cm}
  \centering % Center the figure within the minipage
  \includegraphics[width=0.90\linewidth, keepaspectratio]{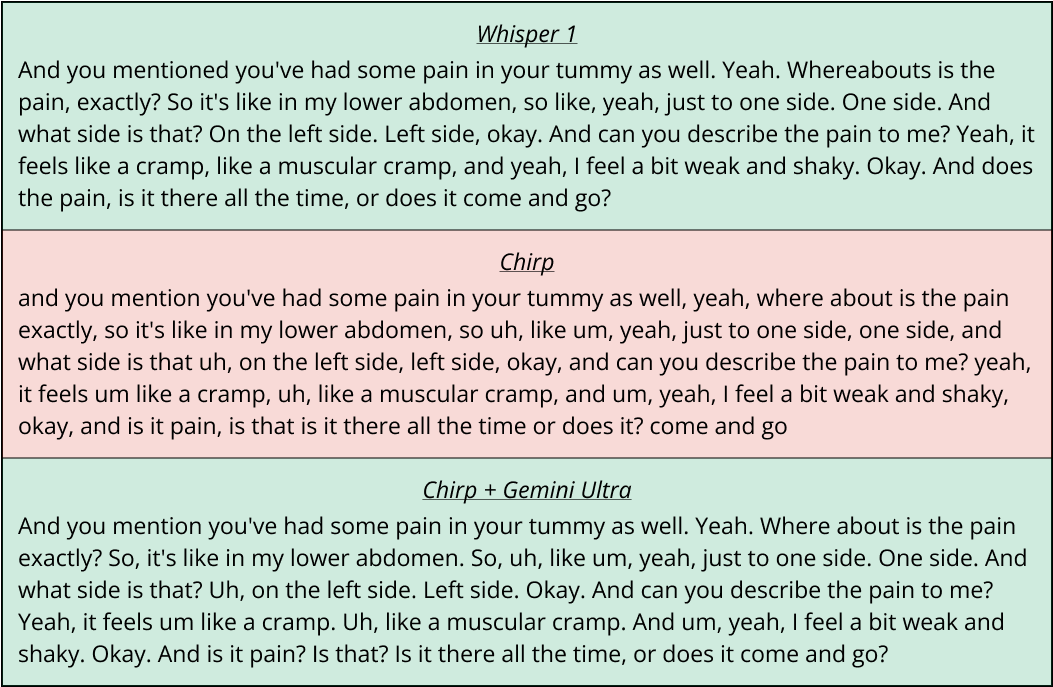} % Adjust the width as needed
  \captionof{figure}{Comparative snippet illustrating transcript quality with Whisper 1 (top) showing high-quality punctuation, Chirp (middle) showing adversarial punctuation, and enhanced output by Gemini Ultra (bottom) with improved punctuation.}
  \label{fig:adversarial_punctuation}
\end{minipage}
\vspace{1em}

Diving into the impact of punctuation on diarization accuracy, our analysis revealed a notable discrepancy in performance when applying CoT iteration to different ASR outputs. We observed that LLM-ASR combinations involving Whisper 1 more consistently achieved state-of-the-art performance and outperformed ASR benchmarks in diarization tasks, while those paired with Chirp performed competitively but lagged by 2 to 4 basis points. This differential in diarization performance is clarified by the comparative punctuation quality of the outputs of each ASR system: Whisper 1 inherently provides an orthographic transcription style, replete with well-placed punctuation, whereas Chirp tends towards a normalized style, characterized by run-on sentences and often inadequate punctuation. This difference in transcription styles is illustrated in Figure 9, which showcases the superior punctuation quality of Whisper 1 compared to the adversarial punctuation patterns found in Chirp's output.\\
\indent It is important to note the observed discrepancy in diarization performance is not simply a matter of comparative transcription accuracy, but rather a fundamental distinction in the training approach of each ASR system. Past research has shown that ASR models can be trained for either orthographic outputs, using datasets that include proper punctuation and casing, or for lower word error rates by normalizing datasets to exclude casing and punctuation \citep{Radford2022robust, Gandhi2022esb}. The latter approach simplifies the task for the ASR model, as it does not have to differentiate between upper and lower case characters or infer punctuation solely from audio cues, yielding naturally lower WERs. In our study, Chirp's results align with these findings, exhibiting a lower general WER in comparison to Whisper 1—a point that we will explore in greater depth later on. This juxtaposition of a lower general WER but less-than-optimal diarization accuracy highlights the challenges that LLMs encounter when diarizing transcripts with suboptimal punctuation.\\
\indent To mitigate the effects of adversarial punctuation, we incorporated a punctuation enhancement step at the beginning of our CoT workflow. This adjustment proved to be effective—as demonstrated by Gemini Ultra's enhanced output in Figure 9—and helped to equalize diarization outcomes. Nevertheless, the results indicate that initial punctuation quality continued to influence diarization outcomes, reaffirming its importance.

\subsection{Medical Concept WER}
Diving into the MC-WER results, we performed a comprehensive comparison of ASR benchmarks—including Amazon Transcribe Medical, Whisper 1, Chirp, Deepgram Nova 2, GCMC, and Soniox—with LLM-ASR combinations. The LLM-ASR combinations involved CoT iterations against Chirp's, Whisper 1's, and GCMC's outputs both in discrete and aggregated 10-line chunks and in their entirety all at once to assess the effectiveness of LLMs across varying input context window sizes. The findings (as detailed in Table 4 and illustrated in Figure 10)  revealed a nuanced interplay between the efficacy of each ASR system and each LLM's proficiency in identifying and correcting errors related to medical concepts.

In both the 10-line chunk and all at once experiments, Whisper 1—either independently or when paired with LLMs like GPT-4 and Gemini Ultra—consistently exhibited the lowest MC-WER, indicating a superior baseline in capturing medical concepts accurately. Notably, the top GPT-4 and Whisper 1 pairing corrected 11 medical concept errors over the ASR baseline in the 10-line chunk experiments. The improvement was less pronounced with the second best Gemini Ultra and Whisper 1 pairing in the same experimental conditions, achieving a reduction of just one error across all transcripts. The modest improvements in total errors made by the LLMs against Whisper 1's outputs — compared to the more drastic improvements seen with Chirp — can largely be traced back to the specific error patterns of Whisper 1 (illustrated in Figure 11).

\begin{minipage}{\dimexpr\textwidth}
\fontsize{8pt}{11pt}\selectfont % Set the font size to 9pt and the line spacing to 11pt
\centering
\begin{tabular}{| l r | r | r | r |}
\hline
                 Model(s) &                   Experiment &            MC-WER &  Error Total & $\Delta$ \\
\hline
        GPT-4 + Whisper 1 &                 Diarization  &   6.25\% ± 4.00 &          312 &         \negative{-11} \\
 Gemini Ultra + Whisper 1 & Diarization + Correction  &   6.44\% ± 4.47 &          322 &          \negative{-1} \\
                Whisper 1 &                          ASR &   6.51\% ± 4.06 &          323 &            \\
   Gemini Pro + Whisper 1 & Diarization + Correction  &   6.54\% ± 4.05 &          322 &          \negative{-1} \\
                   Soniox &                          ASR &   7.33\% ± 4.10 &          372 &             \\
Amazon Transcribe Medical &                          ASR &   8.68\% ± 4.46 &          441 &             \\
     Gemini Ultra + Chirp & Diarization + Correction  &   9.76\% ± 4.56 &          499 &        \negative{-486} \\
          Deepgram Nova 2 &                          ASR &   9.97\% ± 5.61 &          495 &             \\
            GPT-4 + Chirp &                 Diarization  &  10.15\% ± 4.45 &          517 &        \negative{-468} \\
       Gemini Pro + Chirp &                  Correction  &  11.92\% ± 5.39 &          615 &        \negative{-370} \\
                    Chirp &                          ASR &  19.34\% ± 7.83 &          985 &             \\
      Gemini Ultra + GCMC &                  Correction  &  23.11\% ± 9.81 &         1170 &         \negative{-66} \\
    text-bison-32k + GCMC &                  Correction  & 24.43\% ± 10.18 &         1220 &         \negative{-16} \\
             GPT-4 + GCMC &                 Diarization  & 24.61\% ± 10.15 &         1229 &          \negative{-7} \\
        Gemini Pro + GCMC &                 Diarization  & 24.65\% ± 10.78 &         1220 &         \negative{-16} \\
                     GCMC &                          ASR & 25.00\% ± 10.69 &         1236 &             \\
         Claude V2 + GCMC &                 Diarization  & 25.19\% ± 10.43 &         1241 &          \positive{+5} \\
\hline
\end{tabular}
\captionof{table}{MC-WER performance of top LLM-ASR pairings and ASR benchmarks in the 10-line chunk experiments, including standard deviation, total medical concept error count, and the delta ($\Delta$) representing the change from ASR baseline error count.}
\label{tab:mc_wer_table-10_chunks}
\vspace{2em}
\includegraphics[width=\linewidth, keepaspectratio]{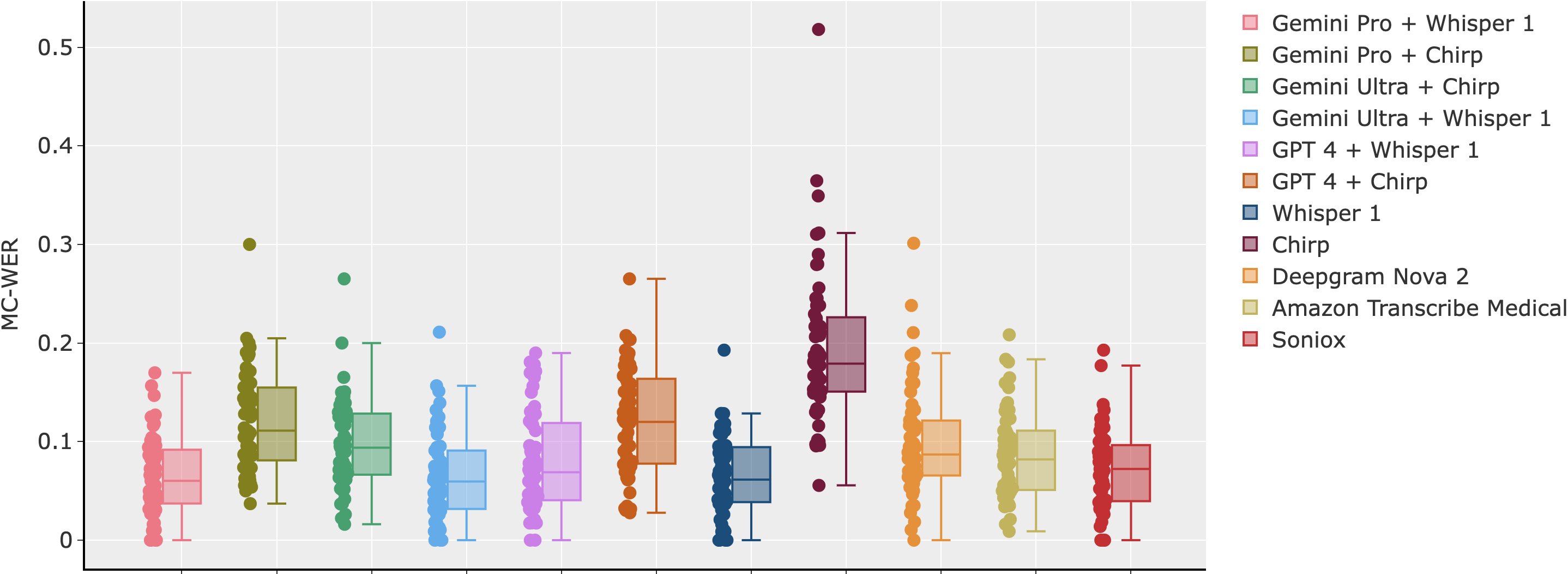} % Adjust the width as needed
\captionof{figure}{Box plot distribution of MC-WER for top-performing LLM and STT pairings when processing entire transcripts in 10-line chunks, including ASR benchmarks.}
\label{fig:mc_wer_box_plots-10_chunks}
\end{minipage}

Whisper 1 transcribed on average 1.72 substitutions, 1.84 deletions, and 2.09 insertions of medical concept errors per transcript. We observed that, across all ASR systems and pairings, insertions and deletions were more intransigent to correction compared to substitutions which were more readily rectified. A key factor in the only modest improvement observed with Whisper 1's outputs was the contextual plausibility of substitution errors in its output; often, these errors remained contextually coherent (e.g., "antibodies" instead of "antibiotics"). Relatedly, many of Whisper 1’s substitutions were rooted in lemmatization differences (e.g, "achy pain" instead of "aching pain"), with such discrepancies contributing to 38 of the 312 identified medical concept errors.\\
\indent In stark contrast, Chirp’s baseline MC-WER of 19.34\% was dramatically reduced to 9.76\% in its most effective LLM-ASR pairing with Gemini Ultra, effectively slashing the total error count from the ASR baseline of 985 errors to just 499. Chirp transcribed on average 6.71 substitutions, 7.61 deletions, and 2.89 insertions of medical concept errors per transcript. When paired with Gemini Ultra, this improved to 2.31 substitutions, 3.77 deletions, and 2.67 insertions of medical concept errors on average per transcript. A key factor in the dramatic improvement observed with Chirp's outputs was the fact its substitutions were often phonetic misinterpretations that diverged significantly from accurate medical and lexical terminology. Unlike Whisper, which generated contextually coherent and lexically accurate hypotheses, Chirp’s errors tended to be more detectable and phonetically distant from the actual terms, such as "emolians" for "emollients" or "exthema" for "eczema." This characteristic of Chirp's errors made them more conspicuous and, consequently, easier for LLMs to correct, leading to the substantial improvements in MC-WER.\\
\indent GCMC presented a distinctive error profile compared to other ASR benchmarks (illustrated in Figure 13). In its most effective pairing with Gemini Ultra, GCMC exhibited a moderate improvement, reducing the total error count by 66 from its ASR baseline. GCMC transcribed on average 8.32 insertions, 8.12 deletions, and 5.24 substitutions per transcript. This error pattern was characterized by a higher frequency of insertions and deletions, often resulting from its transcription of semantically infeasible statements—a key factor that contributed to the moderate improvements observed, unlike the more dramatic results seen with Chirp.\\
\indent For instance, GCMC produced transcriptions like “And was your bomits on nasal crap, but was it just normal food color?” and “Do you mean you gained to 200 more often,” which diverged significantly in coherence from the actual phrases in the ground truth, “And was your vomit, I know it's not a nice thing to talk about, but was it just normal food color?” and “Do you mean you have gone to the bathroom more often?” While LLMs made commendable efforts to reformulate these incoherent statements into more semantically aligned alternatives, these enhancements did not translate effectively into MC-WER improvements due to the metric’s rigid definition. For example, an actual transformation of “Do you mean you gained to 200 more often” to “Do you mean you have gone to the \textit{bathroom} more often?” made the transcript more coherent and closer to the ground truth “Do you mean you have gone to the \textit{toilet} more often?” Yet, such substantial improvements in semantic coherence were not adequately captured by MC-WER, underscoring the limitations inherent in the WER assessment for capturing nuances in semantic coherence.

This challenge was compounded by the nature of substitutions in GCMC's outputs, which exhibited higher character-level error rates compared to Chirp's phonetic misinterpretations with lower character-level error rates. The nature and severity of errors in GCMC's outputs, particularly the frequency of semantically infeasible statements and the higher character error rates in substitutions, posed significant hurdles for LLMs in achieving substantial improvements. Consequently, the moderate improvements from the ASR baseline observed with GCMC highlight the obscured benefits of semantic enhancements and overall complexity of the task.

Moving on, the CoT workflow dynamics were intriguing, with 8 out of the 11 top LLM-ASR combinations including a diarization instruction, and 5 out of these 11 only involving a diarization instruction. This suggests that most LLMs, particularly GPT-4 based on the results, intuitively understand correct medical terminology and silently correct errors even when tasked with a different or broader objective like diarization. This contrasts with explicit correction instruction, where it was observed that some LLMs, particularly GPT-4, begin to overcorrect, modifying both conspicuous errors and non-errors. Conversely, some models like Gemini Ultra and Gemini Pro saw improved outcomes when diarization was explicitly coupled with correction, suggesting that these models benefitted from a two-phase approach to error amendment.

\begin{minipage}{\dimexpr\textwidth}
\begin{adjustwidth}{-2cm}{-2cm}
\includegraphics[width=0.90\linewidth, keepaspectratio]{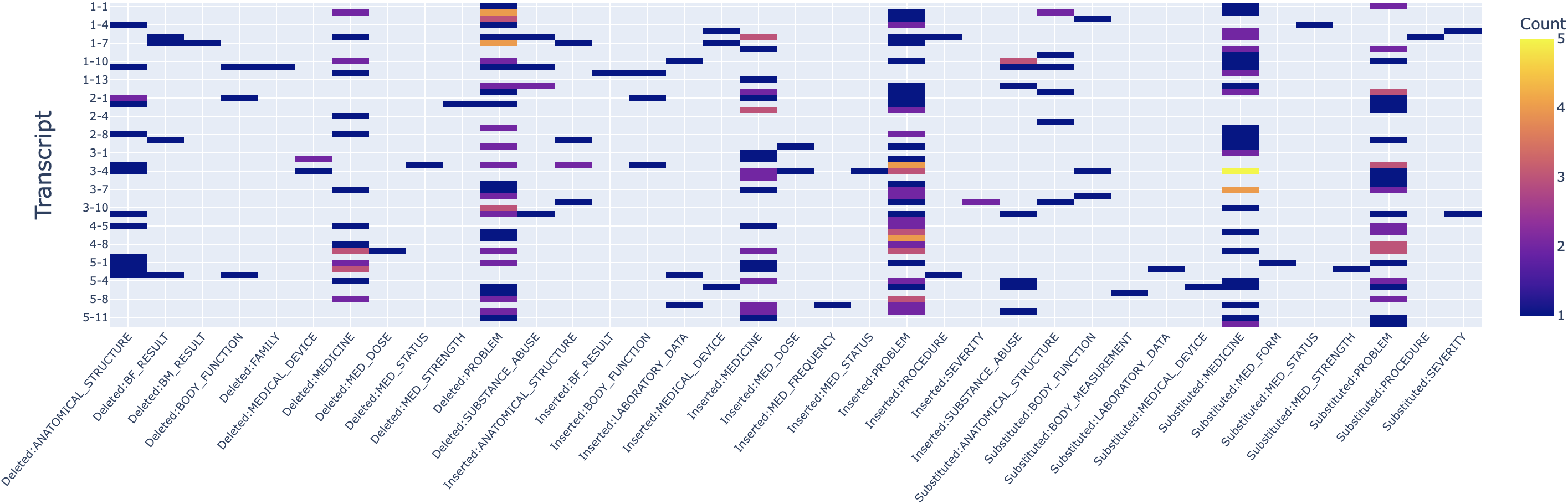}
\end{adjustwidth}
\captionof{figure}{ Heatmap of medical concept errors in transcription by Whisper 1, categorized by type and frequency.}
\label{fig:whisper_mc_wer_heatmap}
\vspace{2em}
\begin{adjustwidth}{-2cm}{-2cm}
\includegraphics[width=0.90\linewidth, keepaspectratio]{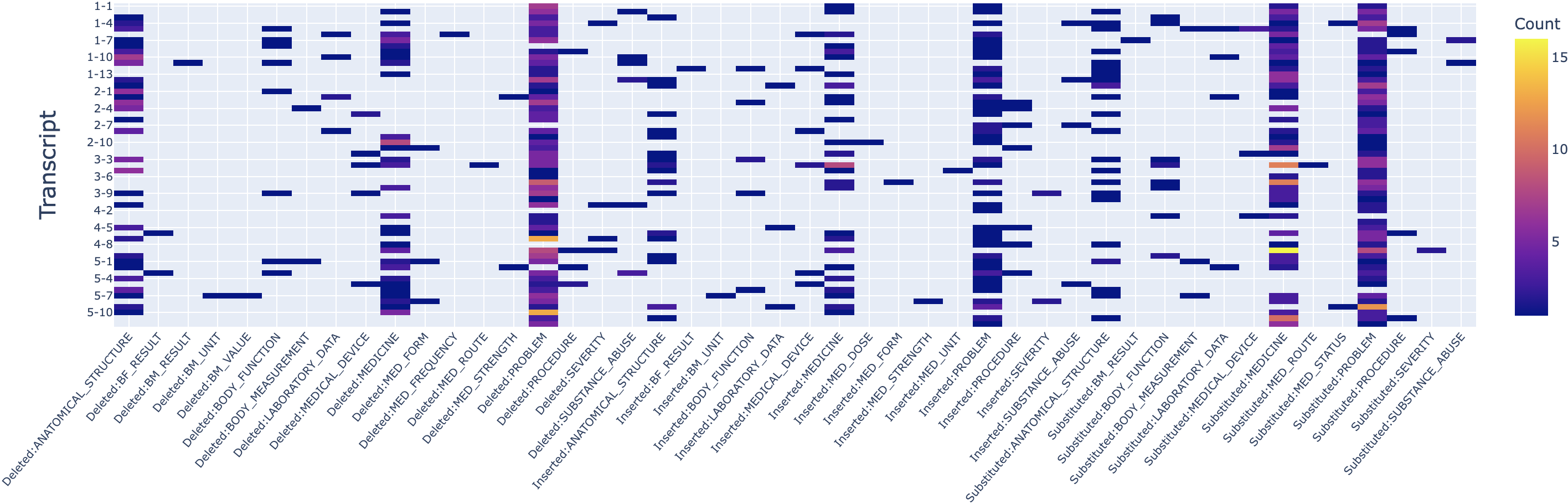}
\end{adjustwidth}
\captionof{figure}{ Heatmap of medical concept errors in transcription by Chirp, categorized by type and frequency.}
\label{fig:chirp_mc_wer_heatmap}
\vspace{2em}
\begin{adjustwidth}{-2cm}{-2cm}
\includegraphics[width=0.90\linewidth, keepaspectratio]{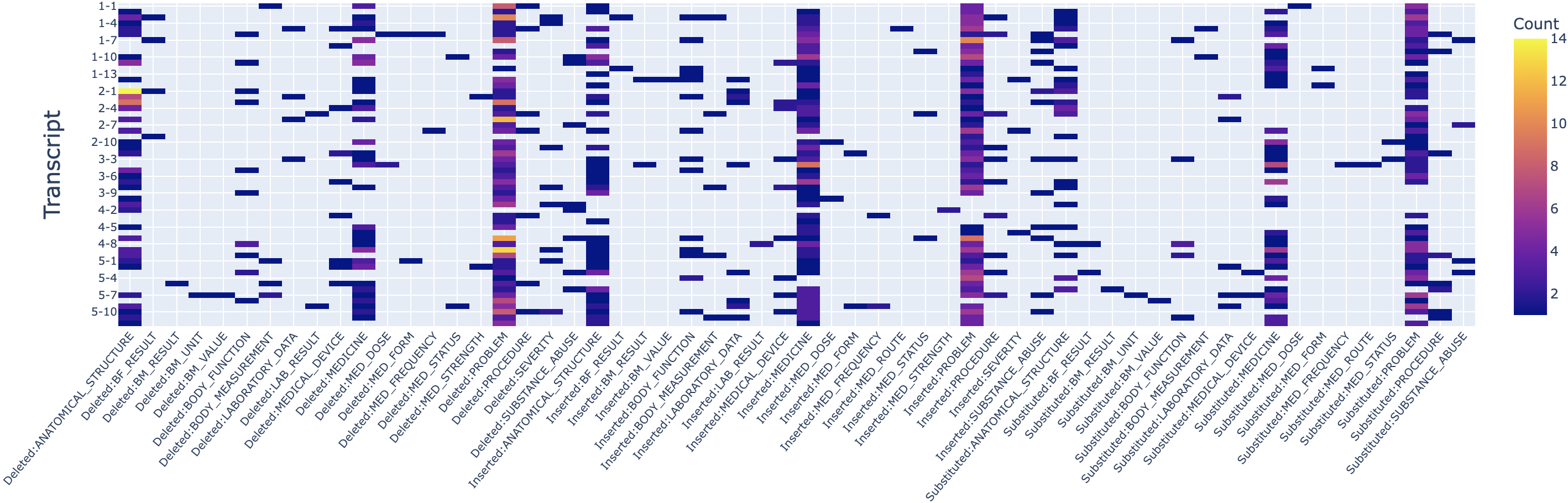}
\end{adjustwidth}
\captionof{figure}{ Heatmap of medical concept errors in transcription by GCMC, categorized by type and frequency.}
\label{fig:gcmc_mc_wer_heatmap}
\end{minipage}
\vspace{2em}

\begin{minipage}{\dimexpr\textwidth} % Use the full width of the page
\begin{adjustwidth}{-0.2cm}{0cm}
\setlength{\parindent}{1em}
The heatmaps depicted in Figures 11, 12, and 13 provide a visual fingerprint of each ASR system's propensity and susceptibility to specific medical concept errors. For instance, Chirp's proneness to omitting anatomical structures and the universal vulnerability to substituting medication terms reveal the subtle complexities of transcribing medical terminology. These insights offer a clear pathway for targeted fine-tuning using techniques like LoRA to refine ASR performance, reinforcing the promising potential of LLM-ASR combinations in medical transcription.
\FloatBarrier

\subsection{General WER}
\FloatBarrier
Moving upstream to the more broadly defined general WER results, we performed a comprehensive comparison of ASR benchmarks—including Amazon Transcribe Medical, Whisper 1, Chirp, Deepgram Nova 2, GCMC, and Soniox—with LLM-ASR combinations. The LLM-ASR combinations involved CoT iterations against Chirp's, Whisper 1's, and GCMC's outputs both in discrete and aggregated 10-line chunks and in their entirety all at once to assess the effectiveness of LLMs across varying input context window sizes.\\
\indent The results (as detailed in Table 5 and illustrated in Figure 14) for the 10-line chunk experiments revealed that most ASR benchmarks and LLM-ASR combinations clustered around 12\% accuracy, with Amazon Transcribe Medical achieving the best all around accuracy of 10.12\%. In the cases of Chirp and GCMC, the top pairings with Gemini Ultra and Gemini Pro yielded modest improvements in general WER accuracy. On the other hand, for Whisper 1, its top pairings with Gemini Ultra and Gemini Pro yielded slight reductions in general accuracy.\\
\indent The outcomes in experiments processing transcripts in their entirety all at once followed a similar pattern for Chirp and Whisper 1, but reversed direction for GCMC (as detailed in Table 6 and illustrated in Figure 15). We suspect this reversal is due to heightened complexity in the processing of GCMC's outputs—which have a greater proportion of semantically infeasible statements, insertions, and deletions—across an entire transcript in a single inference step. The decreased performance in general WER suggests a susceptibility to cognitive overload with very large context windows \citep{Liu2023lost, Xu2023cognitive}. Owing to a general susceptibility to cognitive overload, the LLMs demonstrated no marked improvement when provided with larger context.

\vspace{1em}
\centering
\fontsize{8pt}{11pt}\selectfont % Set the font size to 9pt and the line spacing to 11pt
\setlength{\tabcolsep}{8pt} % Adjust the value to your preference
\begin{tabular}{| l r r | r |}
\hline
           LLM &                       STT &                   Method &         WER \\
\hline
            -- & Amazon Transcribe Medical &                      ASR & 10.12\% ± 2.38 \\
  Gemini Ultra &                     Chirp &               Correction & 11.94\% ± 2.65 \\
    Gemini Pro &                     Chirp &               Correction & 11.97\% ± 2.71 \\
            -- &                     Chirp &                      ASR & 12.06\% ± 2.74 \\
            -- &                 Whisper 1 &                      ASR & 12.14\% ± 3.40 \\
            -- &                    Soniox &                      ASR & 12.16\% ± 3.02 \\
            -- &           Deepgram Nova 2 &                      ASR & 12.25\% ± 3.85 \\
         GPT-4 &                     Chirp &              Diarization & 12.34\% ± 2.83 \\
    Gemini Pro &                 Whisper 1 &              Diarization & 12.42\% ± 3.39 \\
  Gemini Ultra &                 Whisper 1 &              Diarization & 12.42\% ± 3.37 \\
         GPT-4 &                 Whisper 1 &              Diarization & 12.60\% ± 3.27 \\
  Gemini Ultra &                      GCMC &               Correction & 23.00\% ± 5.17 \\
    Gemini Pro &                      GCMC & Diarization + Correction & 23.38\% ± 5.19 \\
            -- &                      GCMC &                      ASR & 23.44\% ± 5.15 \\
text-bison-32k &                      GCMC &              Diarization & 23.49\% ± 5.19 \\
         GPT-4 &                      GCMC &              Diarization & 23.53\% ± 4.98 \\
     Claude V2 &                      GCMC &              Diarization & 23.83\% ± 5.49 \\
\hline
\end{tabular}

\captionof{table}{Optimal Pairings of LLMs and ASR Systems for General WER in experiments processing entire transcripts in 10-line chunks.}
\label{tab:general_wer_accuracy_table-10_chunks}
\end{adjustwidth}
\end{minipage}

\begin{minipage}{\dimexpr\textwidth-0.5cm} % Use the full width of the page
\includegraphics[width=1\linewidth]{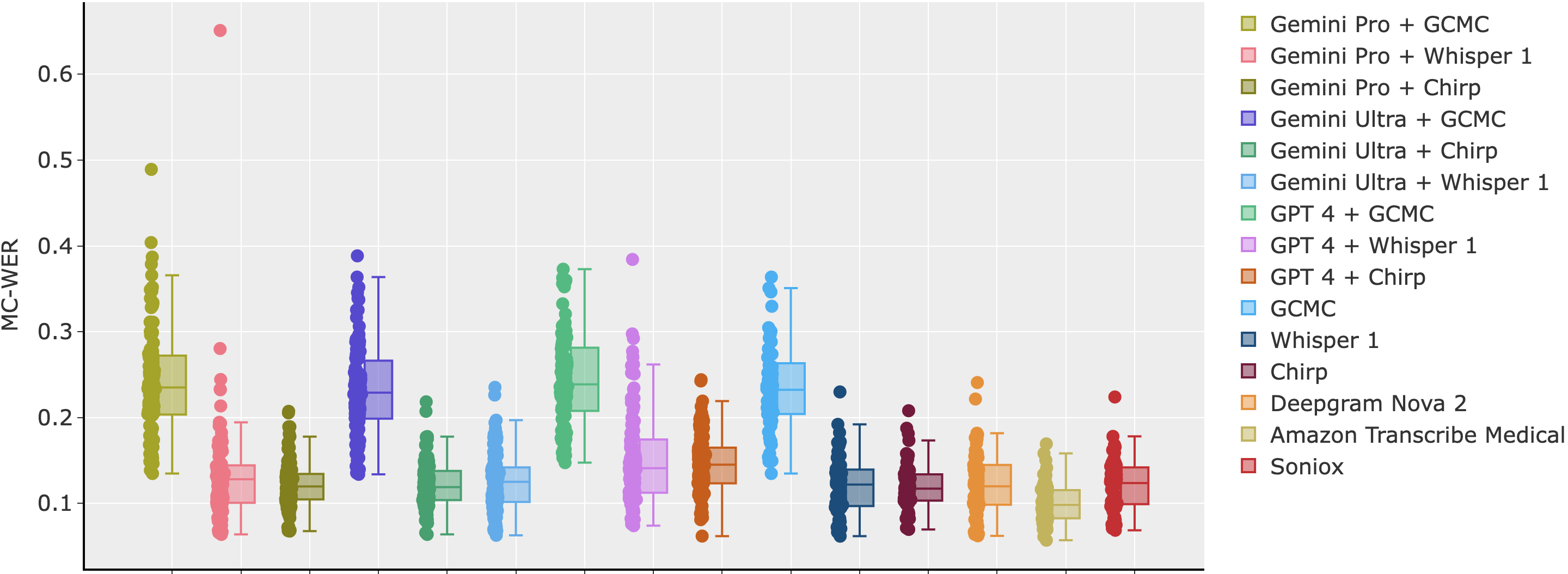} % Adjust the width as needed
\captionof{figure}{Box plot distribution of General WER for top-performing LLM and STT pairings when processing entire transcripts in 10-line chunks.}
\label{fig:general_wer_accuracy_box_plots-10_chunks}
\end{minipage}
\vspace{1.2em}

\begin{minipage}{\dimexpr\textwidth-0.5cm} % Use the full width of the page
\centering
\fontsize{8pt}{11pt}\selectfont % Set the font size to 9pt and the line spacing to 11pt
\setlength{\tabcolsep}{8pt} % Adjust the value to your preference
\begin{tabular}{| l r r | r |}
\hline
           LLM &                       STT &                   Method &          WER \\
\hline
            -- & Amazon Transcribe Medical &                      ASR & 10.12\% ± 2.38 \\
    Gemini Pro &                     Chirp & Diarization + Correction & 12.03\% ± 3.19 \\
            -- &                     Chirp &                      ASR & 12.06\% ± 2.74 \\
text-bison-32k &                     Chirp &              Diarization & 12.13\% ± 3.32 \\
            -- &                 Whisper 1 &                      ASR & 12.14\% ± 3.40 \\
            -- &                    Soniox &                      ASR & 12.16\% ± 3.02 \\
     Claude V2 &                     Chirp &              Diarization & 12.21\% ± 3.06 \\
            -- &           Deepgram Nova 2 &                      ASR & 12.25\% ± 3.85 \\
    Gemini Pro &                 Whisper 1 &              Diarization & 12.38\% ± 3.71 \\
text-bison-32k &                 Whisper 1 &              Diarization & 12.57\% ± 3.84 \\
     Claude V2 &                 Whisper 1 &              Diarization & 12.80\% ± 5.00 \\
            -- &                      GCMC &                      ASR & 23.44\% ± 5.15 \\
    Gemini Pro &                      GCMC &              Diarization & 23.52\% ± 5.33 \\
     Claude V2 &                      GCMC &              Diarization & 24.51\% ± 5.92 \\
text-bison-32k &                      GCMC &              Diarization & 24.71\% ± 5.63 \\
\hline
\end{tabular}

\captionof{table}{Optimal Pairings of LLMs and ASR Systems for General WER in experiments processing entire transcripts all at once.}
\label{tab:general_wer_accuracy_table-all_at_once}
\vspace{2em}

\includegraphics[width=1\linewidth]{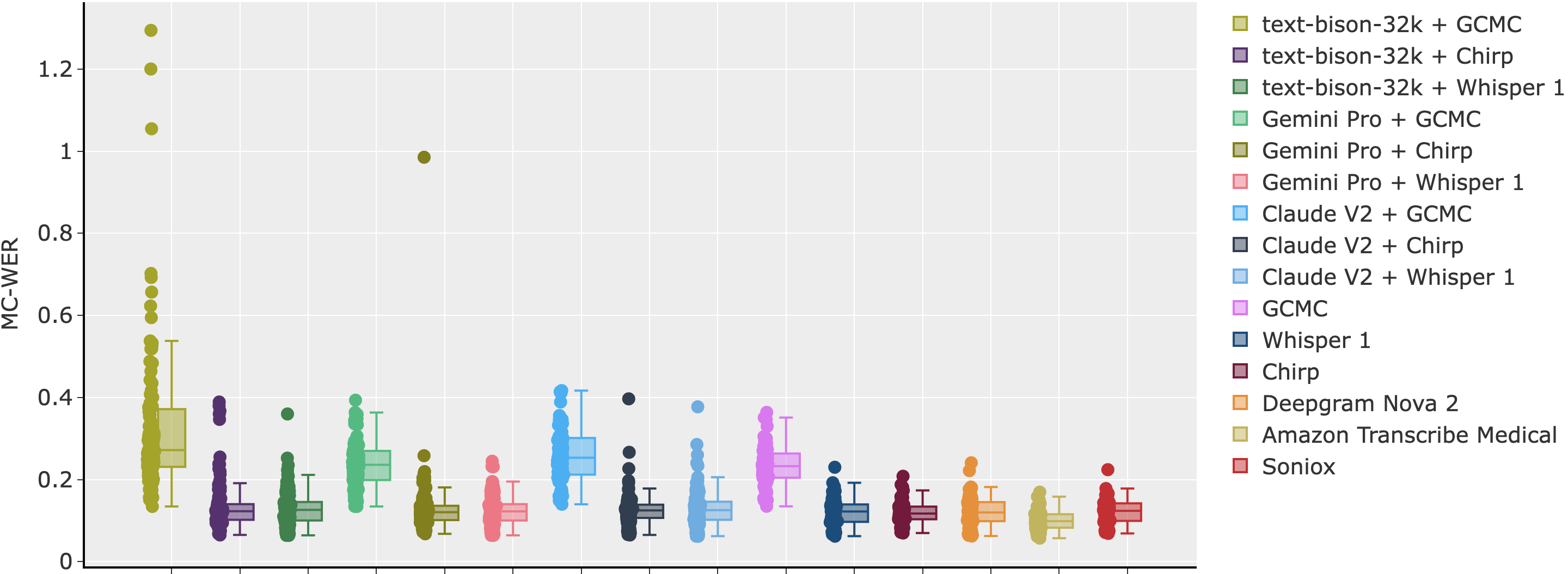} % Adjust the width as needed
\captionof{figure}{Box plot distribution of General WER for top-performing LLM and STT pairings when processing entire transcripts all at once.}
\label{fig:general_wer_accuracy_box_plots-all_at_once}
\end{minipage}
\vspace{1em}

Similar CoT workflow dynamics to those seen in the MC-WER outcomes were observed, with 17 out of the 20 top LLM-ASR combinations including a diarization instruction, and 15 out of these 17 only involving a diarization instruction. This finding reaffirms the notion that LLMs are susceptible to over-correction when explicitly instructed to correct. It also indicates that optimal performance in correction tasks favors the silent amendment of conspicuous errors.

\subsection{Semantic Similarity}
\FloatBarrier
Shifting our attention from word-for-word correction dynamics to the subtleties of semantic textual similarity, we conducted a comparison of ASR benchmarks—including Amazon Transcribe Medical, Whisper 1, Chirp, Deepgram Nova 2, GCMC, and Soniox—with LLM-ASR combinations. The LLM-ASR combinations involved CoT iterations against Chirp's, Whisper 1's, and GCMC's outputs both in discrete and aggregated 10-line chunks and in their entirety all at once to assess the effectiveness of LLMs across varying input context window sizes. This approach allowed us to delve deeper into the nuanced improvements made by the LLMs in semantic coherence and similarity, providing a more granular view of enhancements beyond word-for-word accuracy.\\
\indent In experiments processing transcripts in 10-line chunks and in their entirety all at once, the results consistently indicated that the top LLM-ASR combinations enhanced semantic similarity with the reference transcripts (as detailed in Tables 7 through 10). Notably, this improvement was not restricted to a single ASR system; it was a universal trend seen across Chirp, Whisper 1, and GCMC, and was consistent irrespective of the embedding models applied, which included Google Cloud's PaLM Gecko, OpenAI's Ada, BERT, and RoBERTa. These findings reveal and underscore the capacity of LLMs to refine and enrich the semantic content of transcriptions.

Relatedly, these findings are particularly relevant to the application of LLMs in tasks like Retrieval Augmented Generation (RAG), where the retrieval of accurate and contextually relevant information is paramount \citep{Manathunga2023retrieval, Zhao2023retrieving}. In the medical domain, where precision and recall of information can be critical, the integration of LLMs into the post-processing and summarization of medical records can lead to substantial enhancements in the recall accuracy and quality of information retrieved.
\vspace{1.2em}

\begin{minipage}{\dimexpr\textwidth-0.5cm} % Use the full width of the page
\centering
\fontsize{8pt}{11pt}\selectfont % Set the font size to 9pt and the line spacing to 11pt
\setlength{\tabcolsep}{8pt} % Adjust the value to your preference
\begin{tabular}{| l r r | r |}
\hline
           LLM &                       STT &                   Method &        Cosine Similarity \\
\hline
    Gemini Pro &                 Whisper 1 & Diarization + Correction & 97.38\% ± 6.00 \\
            -- &                 Whisper 1 &                      ASR & 97.37\% ± 5.87 \\
  Gemini Ultra &                 Whisper 1 &               Correction & 97.35\% ± 5.92 \\
            -- & Amazon Transcribe Medical &                      ASR & 97.28\% ± 6.06 \\
            -- &                    Soniox &                      ASR & 97.11\% ± 6.23 \\
            -- &           Deepgram Nova 2 &                      ASR & 96.88\% ± 6.45 \\
  Gemini Ultra &                     Chirp & Diarization + Correction & 96.72\% ± 6.56 \\
    Gemini Pro &                     Chirp & Diarization + Correction & 96.55\% ± 6.63 \\
            -- &                     Chirp &                      ASR & 95.87\% ± 7.13 \\
  Gemini Ultra &                      GCMC &               Correction & 92.81\% ± 9.66 \\
     Claude V2 &                      GCMC & Diarization + Correction & 92.53\% ± 9.80 \\
    Gemini Pro &                      GCMC & Diarization + Correction & 92.44\% ± 9.74 \\
text-bison-32k &                      GCMC & Diarization + Correction & 92.35\% ± 9.74 \\
            -- &                      GCMC &                      ASR & 92.02\% ± 9.89 \\
\hline
\end{tabular}
\captionof{table}{Optimal Pairings of LLMs and ASR Systems for cosine similarity with reference transcript with Google Cloud's PaLM Gecko embeddings in experiments processing entire transcripts in 10-line chunks.}
\label{tab:semantic_similarity_palm_gecko_table-10_chunks}
\end{minipage}

\begin{minipage}{\dimexpr\textwidth-0.5cm} % Use the full width of the page
\centering
\fontsize{8pt}{11pt}\selectfont % Set the font size to 9pt and the line spacing to 11pt
\setlength{\tabcolsep}{8pt} % Adjust the value to your preference
\begin{tabular}{| l r r | r |}
\hline
           LLM &                       STT &                   Method &        Cosine Similarity \\
\hline
    Gemini Pro &                 Whisper 1 & Diarization + Correction & 98.41\% ± 3.13 \\
            -- &                 Whisper 1 &                      ASR & 98.39\% ± 3.06 \\
            -- & Amazon Transcribe Medical &                      ASR & 98.39\% ± 3.10 \\
  Gemini Ultra &                 Whisper 1 &               Correction & 98.38\% ± 3.08 \\
            -- &                    Soniox &                      ASR & 98.26\% ± 3.21 \\
            -- &           Deepgram Nova 2 &                      ASR & 98.20\% ± 3.26 \\
  Gemini Ultra &                     Chirp &               Correction & 98.07\% ± 3.41 \\
    Gemini Pro &                     Chirp & Diarization + Correction & 97.98\% ± 3.46 \\
            -- &                     Chirp &                      ASR & 97.58\% ± 4.17 \\
  Gemini Ultra &                      GCMC &               Correction & 96.02\% ± 4.86 \\
     Claude V2 &                      GCMC & Diarization + Correction & 95.90\% ± 4.92 \\
    Gemini Pro &                      GCMC & Diarization + Correction & 95.86\% ± 4.87 \\
text-bison-32k &                      GCMC & Diarization + Correction & 95.80\% ± 4.91 \\
            -- &                      GCMC &                      ASR & 95.63\% ± 4.96 \\
\hline
\end{tabular}
\captionof{table}{Optimal Pairings of LLMs and ASR Systems for cosine similarity with reference transcript with OpenAI's Ada embeddings in experiments processing entire transcripts in 10-line chunks.}
\label{tab:semantic_similarity_ada_table-10_chunks}
\vspace{2em}
\centering
\fontsize{8pt}{11pt}\selectfont % Set the font size to 9pt and the line spacing to 11pt
\setlength{\tabcolsep}{8pt} % Adjust the value to your preference
\begin{tabular}{| l r r | r |}
\hline
           LLM &                       STT &                   Method &         Cosine Similarity \\
\hline
            -- & Amazon Transcribe Medical &                      ASR & 98.28\% ± 2.88 \\
    Gemini Pro &                 Whisper 1 & Diarization + Correction & 98.11\% ± 3.06 \\
            -- &                 Whisper 1 &                      ASR & 98.07\% ± 3.06 \\
            -- &           Deepgram Nova 2 &                      ASR & 98.04\% ± 3.02 \\
  Gemini Ultra &                 Whisper 1 &              Diarization & 98.03\% ± 3.08 \\
            -- &                    Soniox &                      ASR & 97.97\% ± 3.10 \\
  Gemini Ultra &                     Chirp &               Correction & 97.96\% ± 3.14 \\
    Gemini Pro &                     Chirp &               Correction & 97.91\% ± 3.16 \\
            -- &                     Chirp &                      ASR & 97.75\% ± 3.30 \\
  Gemini Ultra &                      GCMC &               Correction & 96.42\% ± 4.21 \\
    Gemini Pro &                      GCMC & Diarization + Correction & 96.35\% ± 4.19 \\
     Claude V2 &                      GCMC & Diarization + Correction & 96.34\% ± 4.31 \\
text-bison-32k &                      GCMC & Diarization + Correction & 96.27\% ± 4.26 \\
            -- &                      GCMC &                      ASR & 96.18\% ± 4.29 \\
\hline
\end{tabular}
\captionof{table}{Optimal Pairings of LLMs and ASR Systems for cosine similarity with reference transcript with BERT embeddings in experiments processing entire transcripts in 10-line chunks.}
\label{tab:semantic_similarity_bert)table-10_chunks}
\end{minipage}

\begin{minipage}{\dimexpr\textwidth-0.5cm} % Use the full width of the page
\centering
\fontsize{8pt}{11pt}\selectfont % Set the font size to 9pt and the line spacing to 11pt
\setlength{\tabcolsep}{8pt} % Adjust the value to your preference
\begin{tabular}{| l r r | r |}
\hline
           LLM &                       STT &                   Method &         Cosine Similarity \\
\hline
            -- & Amazon Transcribe Medical &                      ASR & 99.98\% ± 0.03 \\
    Gemini Pro &                 Whisper 1 & Diarization + Correction & 99.98\% ± 0.03 \\
            -- &                 Whisper 1 &                      ASR & 99.98\% ± 0.03 \\
  Gemini Ultra &                 Whisper 1 &               Correction & 99.98\% ± 0.03 \\
            -- &           Deepgram Nova 2 &                      ASR & 99.98\% ± 0.04 \\
            -- &                    Soniox &                      ASR & 99.98\% ± 0.03 \\
  Gemini Ultra &                     Chirp &               Correction & 99.98\% ± 0.04 \\
    Gemini Pro &                     Chirp &               Correction & 99.98\% ± 0.04 \\
            -- &                     Chirp &                      ASR & 99.97\% ± 0.04 \\
  Gemini Ultra &                      GCMC &               Correction & 99.95\% ± 0.05 \\
     Claude V2 &                      GCMC & Diarization + Correction & 99.95\% ± 0.05 \\
    Gemini Pro &                      GCMC & Diarization + Correction & 99.95\% ± 0.05 \\
text-bison-32k &                      GCMC & Diarization + Correction & 99.95\% ± 0.05 \\
            -- &                      GCMC &                      ASR & 99.95\% ± 0.05 \\
\hline
\end{tabular}
\captionof{table}{Optimal Pairings of LLMs and ASR Systems for cosine similarity with reference transcript with RoBERTa embeddings in experiments processing entire transcripts in 10-line chunks.}
\label{tab:semantic_similarity_roberta_table-10_chunks}
\end{minipage}

\subsection{Zero-shot Prompting}
Diving into the zero-shot prompting results, we conducted a comparison of ASR benchmarks—Whisper 1 and GCMC—with various  LLM-ASR combinations. These combinations involved blended diarization and correction instruction against the outputs of Whisper 1 and GCMC, in discrete and aggregated chunks of 5 and 10 lines to assess the effectiveness of LLMs across varying input context window sizes. Notably, in the inferences against GCMC's output, ASR-generated speaker labels were not removed before prompting. This approach facilitated an evaluation of the LLMs' zero-shot diarization capabilities, particularly in enhancing a semi-accurate ASR baseline compared to diarizing entirely from an unlabeled starting point.

Our investigation into general WER improvement revealed that zero-shot prompting could not typically achieve the level of performance exhibited with CoT iteration (as detailed in Table 11 and illustrated in Figure 16). A positive correlation between increased context window size and general WER task performance was evident, with nearly all LLMs exhibiting improved accuracy when the context window was expanded from 5 to 10 lines of dialogue. All LLM-ASR combinations failed to surpass the ASR benchmarks, highlighting the critical role of CoT instruction and the utility of few-shot examples for optimal output. However, a notable exception was the pairing of Gemini Pro and Whisper 1, which approximated its CoT performance with an accuracy of 12.92\% in zero-shot conditions. This suggests that certain LLM configurations may bridge the gap between zero-shot and CoT instruction.

Conversely, it was notable that two of the largest models, GPT-4 and Gemini Ultra, were among the poorest performers in this task. This pattern suggests a heightened susceptibility for over-reasoning and over-correcting in scenarios where instructions are minimal, leading to diminished performance. Such outcomes suggest that smaller, more nimble models may be more adept in zero-shot conditions, where a clear reasoning pathway is not provided for transcription enhancement.

\vspace{1em}
\begin{minipage}{\dimexpr\textwidth-0.5cm} % Use the full width of the page
\centering
\fontsize{8pt}{11pt}\selectfont % Set the font size to 9pt and the line spacing to 11pt
\setlength{\tabcolsep}{8pt} % Adjust the value to your preference
\begin{tabular}{| l r r r | r |}
\hline
           LLM &       STT & Chunk Size &                     Method &          WER \\
\hline
            -- & Whisper 1 &          - &                        ASR & 12.14\% ± 3.40 \\
    Gemini Pro & Whisper 1 &         10 & Diarization + Correction & 12.95\% ± 3.59 \\
text-bison-32k & Whisper 1 &         10 & Diarization + Correction & 14.72\% ± 3.71 \\
    Gemini Pro & Whisper 1 &          5 & Diarization + Correction & 15.57\% ± 4.86 \\
text-bison-32k & Whisper 1 &          5 & Diarization + Correction & 15.61\% ± 3.82 \\
     Claude V2 & Whisper 1 &          5 & Diarization + Correction & 18.31\% ± 4.92 \\
     Claude V2 & Whisper 1 &         10 & Diarization + Correction & 18.94\% ± 5.96 \\
  Gemini Ultra & Whisper 1 &         10 & Diarization + Correction & 19.68\% ± 4.91 \\
  Gemini Ultra & Whisper 1 &          5 & Diarization + Correction & 22.21\% ± 4.76 \\
            -- &      GCMC &          - &                        ASR & 23.44\% ± 5.15 \\
    Gemini Pro &      GCMC &         10 & Diarization + Correction & 23.77\% ± 5.16 \\
text-bison-32k &      GCMC &          5 & Diarization + Correction & 23.93\% ± 5.24 \\
         GPT-4 & Whisper 1 &         10 & Diarization + Correction & 24.40\% ± 5.49 \\
    Gemini Pro &      GCMC &          5 & Diarization + Correction & 24.70\% ± 5.25 \\
     Claude V2 &      GCMC &          5 & Diarization + Correction & 25.78\% ± 6.48 \\
         GPT-4 & Whisper 1 &          5 & Diarization + Correction & 26.21\% ± 4.92 \\
text-bison-32k &      GCMC &         10 & Diarization + Correction & 26.43\% ± 5.41 \\
 LLaMA 2 (70B) &      GCMC &          5 & Diarization + Correction & 27.51\% ± 6.84 \\
  Gemini Ultra &      GCMC &         10 & Diarization + Correction & 28.23\% ± 5.07 \\
  Gemini Ultra &      GCMC &          5 & Diarization + Correction & 29.37\% ± 5.23 \\
         GPT-4 &      GCMC &          5 & Diarization + Correction & 29.44\% ± 5.04 \\
     Claude V2 &      GCMC &         10 & Diarization + Correction & 30.62\% ± 7.55 \\
         GPT 4 &      GCMC &         10 & Diarization + Correction & 31.02\% ± 4.78 \\
\hline
\end{tabular}
\captionof{table}{Optimal Pairings of LLMs and ASR Systems for General WER in zero-shot experiments processing entire transcripts in 5-line or 10-line chunks.}
\label{tab:zero_shot_general_wer_accuracy_table-5,10_chunks}
\end{minipage}
\vspace{2em}

\begin{minipage}{\dimexpr\textwidth-0.5cm}
\includegraphics[width=1\linewidth]{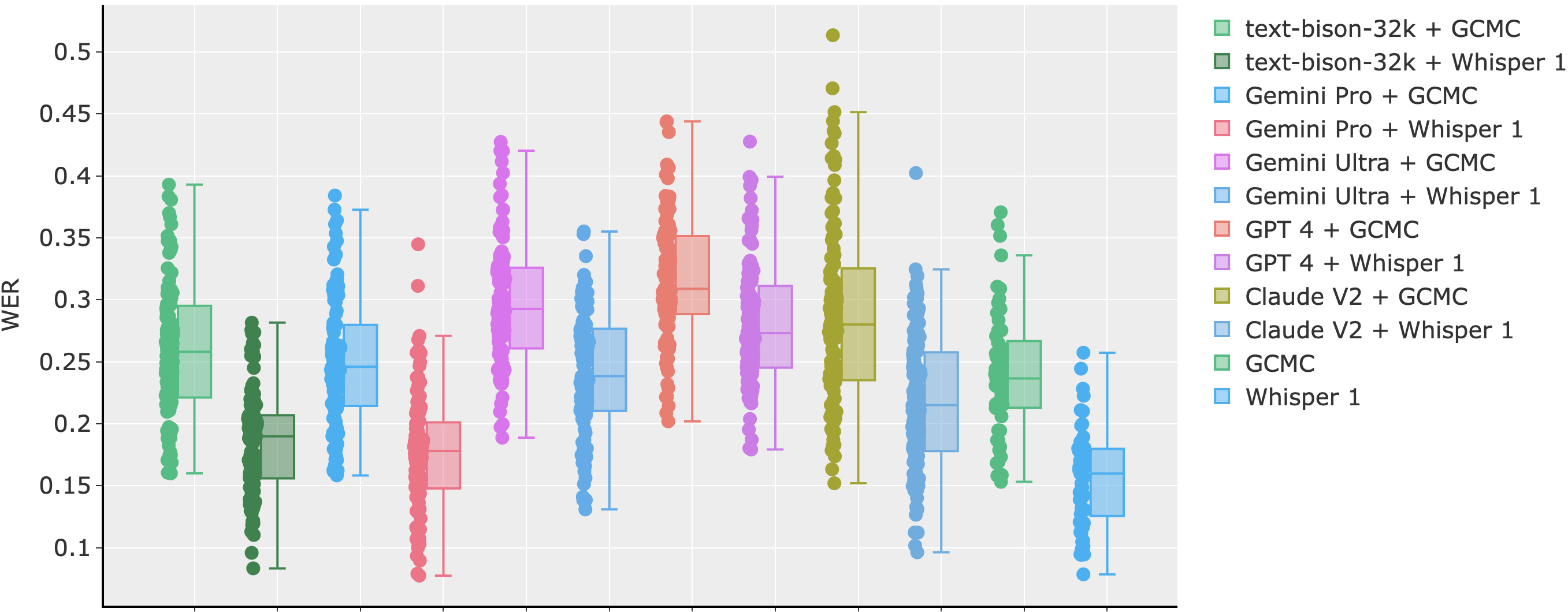} % Adjust the width as needed
\captionof{figure}{Box plot distribution of General WER in zero-shot experiments processing entire transcripts in 5-line or 10-line chunks.}
\label{fig:zero_shot_general_wer_accuracy_box_plots-5,10_chunks}
\end{minipage}

\vspace{2em}
Shifting the focus to diarization-specific metrics, the findings were consistent with those of the general WER task (as detailed in Table 12). Zero-shot outcomes did not come close to those attained with CoT instruction and similar-sized contexts. The best zero-shot result for Doctor-Specific diarization was a D-WER of 30.49\% with the pairing of Gemini Ultra with Whisper 1, and for Patient-Specific diarization, a P-WER of 42.3\% with the pairing of Gemini Ultra and GCMC. Interestingly, the diarization results against GCMC and Whisper 1 outputs were comparable, indictating that the LLMs were as adept at diarizing from scratch as they were at refining semi-accurate ASR outputs.

\begin{minipage}{\dimexpr\textwidth-0.5cm} % Use the full width of the page
\centering
\fontsize{8pt}{11pt}\selectfont % Set the font size to 9pt and the line spacing to 11pt
\setlength{\tabcolsep}{8pt} % Adjust the value to your preference
\begin{tabular}{| l r r r | r |}
\hline
           LLM &       STT & Chunk Size &                     Method &          D-WER \\
\hline
  Gemini Ultra & Whisper 1 &         10 & Diarization + Correction &  30.50\% ± 7.92 \\
     Claude V2 & Whisper 1 &         10 & Diarization + Correction &  31.08\% ± 8.82 \\
text-bison-32k &      GCMC &          5 & Diarization + Correction &  32.43\% ± 8.99 \\
         GPT-4 & Whisper 1 &         10 & Diarization + Correction &  32.53\% ± 8.35 \\
  Gemini Ultra &      GCMC &          5 & Diarization + Correction &  33.06\% ± 8.10 \\
    Gemini Pro &      GCMC &          5 & Diarization + Correction &  33.44\% ± 9.83 \\
  Gemini Ultra &      GCMC &         10 & Diarization + Correction &  33.71\% ± 8.17 \\
    Gemini Pro &      GCMC &         10 & Diarization + Correction & 33.97\% ± 11.00 \\
text-bison-32k & Whisper 1 &         10 & Diarization + Correction &  34.43\% ± 8.62 \\
         GPT-4 &      GCMC &          5 & Diarization + Correction &  34.44\% ± 8.36 \\
     Claude V2 & Whisper 1 &          5 & Diarization + Correction &  34.81\% ± 8.85 \\
    Gemini Pro & Whisper 1 &         10 & Diarization + Correction &  35.09\% ± 8.16 \\
         GPT-4 & Whisper 1 &          5 & Diarization + Correction &  35.42\% ± 8.12 \\
  Gemini Ultra & Whisper 1 &          5 & Diarization + Correction &  35.63\% ± 6.97 \\
text-bison-32k &      GCMC &         10 & Diarization + Correction & 35.88\% ± 10.76 \\
         GPT-4 &      GCMC &         10 & Diarization + Correction &  36.16\% ± 9.43 \\
text-bison-32k & Whisper 1 &          5 & Diarization + Correction &  39.00\% ± 7.12 \\
            -- &      GCMC &        ASR &                        ASR & 39.37\% ± 14.71 \\
    Gemini Pro & Whisper 1 &          5 & Diarization + Correction &  40.55\% ± 7.81 \\
     Claude V2 &      GCMC &          5 & Diarization + Correction & 41.48\% ± 12.36 \\
 LLaMA 2 (70B) &      GCMC &          5 & Diarization + Correction & 43.41\% ± 12.44 \\
     Claude V2 &      GCMC &         10 & Diarization + Correction & 46.41\% ± 13.57 \\
\hline
\end{tabular}
\captionof{table}{Optimal Pairings of LLMs and ASR Systems for Doctor-Specific Diarization in zero-shot experiments processing entire transcripts in 5-line or 10-line chunks.}
\label{tab:zero_shot_diarization_accuracy_doctor_table-5,10_chunks}
\end{minipage}

\begin{minipage}{\dimexpr\textwidth}
\begin{adjustwidth}{-0.4cm}{0cm}
\fontsize{8pt}{11pt}\selectfont % Set the font size to 9pt and the line spacing to 11pt
\centering
\begin{tabular}{| l r | r | r | r | r |}
\hline
                  Model(s) &                               Experiment & Chunk Size &           WER &  Error Total & $\Delta$ \\
\hline
                 Whisper 1 &                                      ASR &            - & 12.14\% ± 3.40 &          322 &             \\
    Gemini Pro + Whisper 1 &              Diarization and Correction  &           10 & 12.95\% ± 3.59 &          371 &         \positive{+49} \\
text-bison-32k + Whisper 1 &              Diarization and Correction  &           10 & 14.72\% ± 3.71 &          469 &        \positive{+147} \\
     Claude V2 + Whisper 1 & Diarization + Correction &            5 & 18.31\% ± 4.92 &          827 &        \positive{+505} \\
  Gemini Ultra + Whisper 1 &              Diarization and Correction  &           10 & 19.68\% ± 4.91 &          604 &        \positive{+282} \\
                      GCMC &                                      ASR &            - & 23.44\% ± 5.15 &         1236 &             \\
         Gemini Pro + GCMC &              Diarization and Correction  &           10 & 23.77\% ± 5.16 &         1252 &         \positive{+16} \\
     text-bison-32k + GCMC & Diarization + Correction &            5 & 23.93\% ± 5.24 &         1430 &        \positive{+194} \\
         GPT-4 + Whisper 1 &              Diarization and Correction  &           10 & 24.40\% ± 5.49 &          927 &        \positive{+605} \\
          Claude V2 + GCMC & Diarization + Correction &            5 & 25.78\% ± 6.48 &         1844 &        \positive{+608} \\
      LLaMA 2 (70B) + GCMC & Diarization + Correction &            5 & 27.51\% ± 6.84 &            1444 &       \positive{+208} \\
       Gemini Ultra + GCMC &              Diarization and Correction  &           10 & 28.23\% ± 5.07 &         1367 &        \positive{+131} \\
              GPT-4 + GCMC & Diarization + Correction &            5 & 29.44\% ± 5.04 &         1513 &        \positive{+277} \\
\hline
\end{tabular}
\end{adjustwidth}
\captionof{table}{MC-WER performance of top LLM-ASR pairings and ASR benchmarks in the zero-shot experiments, including standard deviation, total medical concept error count, and the delta ($\Delta$) representing the change from ASR baseline error count.}
\label{tab:zero_shot_mc_wer_table-5,10_chunks}
\end{minipage}

\vspace{2em}
\begin{minipage}{\dimexpr\textwidth}
\includegraphics[width=\linewidth, keepaspectratio]{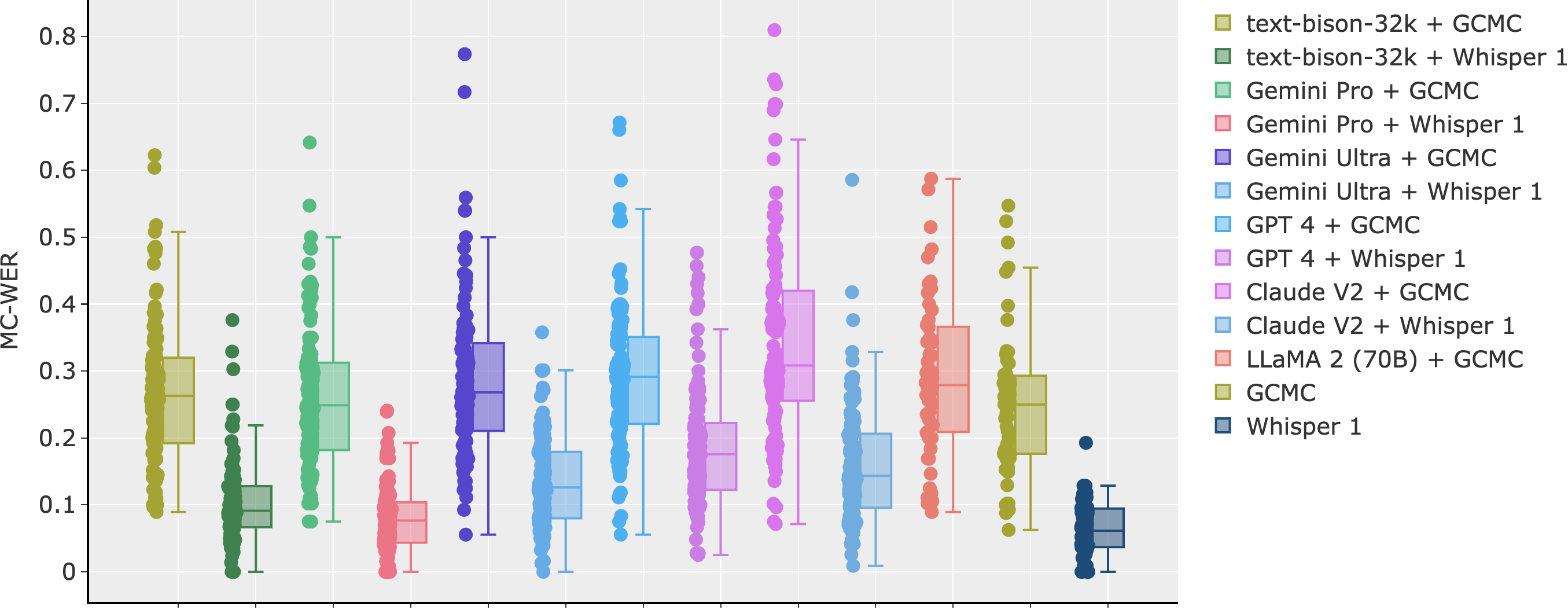} % Adjust the width as needed
\captionof{figure}{Box plot distribution of MC-WER for top-performing LLM and STT pairings in the zero-shot experiments, including ASR benchmarks.}
\label{fig:zero_shot_mc_wer_boxplots-5,10_chunks}
\end{minipage}
\vspace{1em}

Regarding MC-WER results in the zero-shot experiments (as detailed in Table 13 and illustrated in Figure 17), none of the LLM-ASR combinations could replicate the performance achieved with CoT iteration in similar-sized contexts, nor could they reduce the error count from their respective ASR baselines. In the evaluation of more nuanced semantic similarity (as detailed and exemplified in Table 14), this pattern remained consistent; none of the LLM-ASR combinations were able to replicate their performance achieved with CoT iteration in similar-sized contexts, and none surpassed their ASR baselines in terms of semantic similarity. These results underscore again the indispensable role of CoT instruction and few-shot examples in the complex and nuanced task of transcription enhancement. It is evident that while LLMs can outperform ASR benchmarks in diarization tasks, can more accurately capture medical concepts, and can improve the semantic coherence of transcribed dialogues, the success hinges critically on the integration of CoT as an essential component in practical applications.

\begin{minipage}{\dimexpr\textwidth-0.5cm} % Use the full width of the page
\centering
\fontsize{8pt}{11pt}\selectfont % Set the font size to 9pt and the line spacing to 11pt
\setlength{\tabcolsep}{8pt} % Adjust the value to your preference
\begin{tabular}{| l r r r | r |}
\hline
           LLM &       STT & Chunk Size &                     Method &          Cosine Similarity \\
\hline
            -- & Whisper 1 &          - &                        ASR &  99.63\% ± 0.24 \\
     Claude V2 & Whisper 1 &         10 & Diarization + Correction &  98.36\% ± 2.44 \\
            -- &      GCMC &          - &                        ASR &  97.95\% ± 0.90 \\
text-bison-32k & Whisper 1 &         10 & Diarization + Correction &  96.69\% ± 6.50 \\
    Gemini Pro & Whisper 1 &         10 & Diarization + Correction &  96.57\% ± 7.18 \\
    Gemini Pro &     Chirp &         10 & Diarization + Correction &  96.13\% ± 7.65 \\
text-bison-32k &     Chirp &         10 & Diarization + Correction &  96.10\% ± 7.17 \\
  Gemini Ultra & Whisper 1 &         10 & Diarization + Correction &  95.45\% ± 8.25 \\
  Gemini Ultra &     Chirp &         10 & Diarization + Correction &  94.89\% ± 8.70 \\
     Claude V2 &     Chirp &         10 & Diarization + Correction &  94.71\% ± 8.81 \\
     Claude V2 &      GCMC &         10 & Diarization + Correction &  94.38\% ± 5.05 \\
    Gemini Pro &      GCMC &         10 & Diarization + Correction &  92.20\% ± 9.98 \\
text-bison-32k &      GCMC &         10 & Diarization + Correction & 92.02\% ± 10.10 \\
  Gemini Ultra &      GCMC &         10 & Diarization + Correction & 91.87\% ± 10.25 \\
\hline
\end{tabular}
\captionof{table}{Optimal Pairings of LLMs and ASR Systems for cosine similarity with reference transcript with Google Cloud's PaLM Gecko embeddings in zero-shot experiments processing entire transcripts in 10-line chunks.}
\label{tab:zero_shot_semantic_similarity_palm_gecko_table-10_chunks}
\end{minipage}

\section{Limitations}\label{sec:limitations}
Despite the promising results outlined in this study, it is important to acknowledge several limitations that invite further research. To begin, the PriMock57 dataset, while diverse and comprehensive, has limited demographic scope as a single dataset. Expanding the scope of future studies to include a wider range of patient ages, backgrounds, and languages would provide a more robust assessment of generalizability across diverse patient populations.  Similarly, the two-speaker exchanges in the dataset may not fully reflect complex clinical interactions involving multiple participants. Investigating performance in the presence of additional speakers such as nurses and caregivers would shed light on its adaptability to real-world clinical settings.

Furthermore, our study's use of LLMs did not extend to the direct processing of acoustic audio cues. This opens up avenues for future research into multimodal LLMs, which could incorporate audio cues such as tone and pitch into the correction process. 

Lastly, it is important to acknowledge that LLM-mediated transcription enhancement could hide systemic errors within the transcripts. This risk highlights the importance of thoughtful user interface design in real-world applications. Strategies such as integrating confidence scores and linking original audio recordings can enhance transparency and foster trust.  These measures empower critical users, such as medical professionals and transcription reviewers, to more effectively identify and examine potential inaccuracies.

\vspace{-0.7em}
\section{Conclusion}\label{sec:conclusion}
The methodology of our study was carefully crafted to extend beyond mere word-for-word accuracy. It was designed to capture the essence of medical dialogues with an emphasis on accuracy, effectiveness, and \textit{empathy}—a reflection of the human-centric nature of healthcare.  Our comprehensive study illuminates the substantial role LLMs can play in elevating the quality of medical transcripts. This improvement is reflected not just in reduced general WER and MC-WER, but also in improved speaker diarization and increased semantic coherence of ASR-generated transcripts. These advancements can collectively contribute to better clinical workflows and patient outcomes.

Our goal was to test the boundaries of what ASR and LLM technologies can achieve without extensive fine-tuning or modifications, paving the way for their broader application across various environments and tasks. This approach is poised to bridge technological gaps, particularly in resource-limited settings where access to the most advanced on-device ASR and LLM technologies may be out of reach. As we look ahead, our next steps call for a more expansive evaluation of ASR enhancements with LLMs across different languages and medical scenarios. By doing so, we hope to further enhance the practical utility of these models in diverse healthcare environments, ensuring that every medical conversation, regardless of how and where it occurs, meets a high standard of clarity and reliability.

\vspace{-0.7em}
\section{Acknowledgements}\label{sec:acknowledgements}
This work would not have been possible without the help of a number of colleagues, including Dr. Will Morris, Dr. Peter Clardy, Riju Khetarpal, Chris Sakalosky, Hussain Chinoy, Daniel Golden, Timo Kohlberger, Shashir Reddy, and Praney Mittal.

\bibliographystyle{unsrt}
\vspace{-0.5em}
\bibliography{references}

\end{document}